\documentclass[lettersize,journal]{IEEEtran}
\usepackage{amsmath,amsfonts}
\usepackage{algorithmic}
\usepackage{algorithm}
\usepackage{array}
\usepackage{textcomp}
\usepackage{stfloats}
\usepackage{url}
\usepackage{verbatim}
\usepackage{graphicx}
\usepackage{cite}
\usepackage{multirow}
\usepackage{subfigure}
\usepackage{subcaption} % 用于创建子图
\usepackage{amssymb}
\usepackage{amsmath}
\usepackage[table]{xcolor}
\definecolor{grey}{RGB}{128,128,128} % Adjust RGB values as needed
\newcommand\boldp[1]{\vspace{.1cm}\noindent\textbf{#1.}\hspace{.065cm}}

\begin{document}

\title{MCGS: Multiview Consistency Enhancement for Sparse-View 3D Gaussian Radiance Fields}

\author{
Yuru Xiao, Deming Zhai~\IEEEmembership{Member,~IEEE,}, Wenbo Zhao~\IEEEmembership{Member,~IEEE,},  Kui Jiang,~\IEEEmembership{Member,~IEEE,} Junjun Jiang,~\IEEEmembership{Senior Member,~IEEE,} Xianming Liu,~\IEEEmembership{Member,~IEEE}
\IEEEcompsocitemizethanks{
\IEEEcompsocthanksitem  Y. Xiao, D. Zhai, W. Zhao, K. Jiang, J. Jiang, and X. Liu are with the School of Computer Science and Technology, Harbin Institute of Technology, Harbin, China, E-mail: xiaoyuru.30@gmail.com; \{zhaideming, wbzhao, jiangkui, jiangjunjun, csxm\}@hit.edu.cn.
% \IEEEcompsocthanksitem X. Ji is with the Department of Automation, Tsinghua University,
% Beijing, 100084, China, E-mail: xyji@tsinghua.edu.cn.
% \IEEEcompsocthanksitem This work was supported by National Natural Science Foundation of China (xxxx). 
} 

}
\markboth{IEEE Transactions on Pattern Analysis and Machine Intelligence}%
{Xiao \MakeLowercase{\textit{et al.}}: MCGS: Multiview Consistency Enhancement for Sparse-View 3D Gaussian Radiance Fields}



% % column for its text to clear the IEEEpubid mark.

\maketitle

\begin{abstract}
Radiance fields represented by 3D Gaussians excel at synthesizing novel views, offering both high training efficiency and fast rendering. However, with sparse input views, the lack of multi-view consistency constraints results in poorly initialized Gaussians and unreliable heuristics for optimization, leading to suboptimal performance. Existing methods often incorporate depth priors from dense estimation networks but overlook the inherent multi-view consistency in input images. Additionally, they rely on dense initialization, which limits the efficiency of scene representation.
To overcome these challenges, we propose a view synthesis framework based on 3D Gaussian Splatting, named \textbf{MCGS}, enabling photorealistic scene reconstruction from sparse views. The key innovations of MCGS in enhancing multi-view consistency are as follows: {i)  We leverage matching priors from a sparse matcher to initialize Gaussians primarily on textured regions, while low-texture areas are populated with randomly distributed Gaussians. This yields a compact yet sufficient set of initial Gaussians. ii) We propose a multi-view consistency-guided progressive pruning strategy to dynamically eliminate inconsistent Gaussians. This approach confines their optimization to a consistency-constrained space, which ensures robust and coherent scene reconstruction.}
These strategies enhance robustness to sparse views, accelerate rendering, and reduce memory consumption, making MCGS a practical framework for 3D Gaussian Splatting.
\end{abstract}


\section{Introduction}
\IEEEPARstart{N}{ovel} view synthesis (NVS) plays a pivotal role in a wide range of real-world applications, such as VR/AR, robotics, and autonomous driving. Neural Radiance Fields (NeRFs) \cite{mildenhall2021nerf} achieve realistic novel view synthesis from a set of dense input views. However, their performance significantly drops when only sparse views are available, a challenge known as the few-shot neural rendering problem \cite{niemeyer2022regnerf, yang2023freenerf, kim2022infonerf}. Recent studies have made significant strides in addressing this issue, achieving photorealistic novel view synthesis through various techniques such as depth regularization \cite{deng2022depth}, depth distillation \cite{wang2023sparsenerf}, semantic consistency regularization \cite{jain2021putting}, and frequency annealing \cite{yang2023freenerf}. Nonetheless, these methods still face challenges, including costly training and slow rendering speeds due to the limitations of implicit representation. While recent grid-based methods \cite{sun2023vgos, xiao2025spatial} have substantially accelerated training, rendering speeds remain a bottleneck.

3D Gaussian Splatting (3DGS) \cite{kerbl20233d} has employed 3D Gaussian primitives to model the radiance field, achieving success in high-fidelity and real-time novel view synthesis with high training efficiency. However, it still suffers from a performance decline when only sparse input views are available \cite{li2024dngaussian, zhu2024fsgs, zhang2024cor}. This drop is primarily due to its reliance on {sufficient initial point cloud and} multi-view constraint-based optimization, which are less effective under sparse view conditions. The weakened multi-view consistency constraints result in unstable training, especially when the initialized point cloud is extremely sparse.

{Existing methods to address this limitation rely on strong geometric priors, such as depth distillation \cite{zhu2024fsgs} or depth regularization \cite{li2024dngaussian}, derived from pre-trained dense estimation networks. These methods are heavily dependent on the accuracy of the depth estimations and often fail to fully exploit the multi-view consistency present in the input images. Furthermore, their use of a dense initialization from a multi-view stereo (MVS) algorithm results in inefficient scene representation and high memory costs. Consequently, the potential for fully leveraging multi-view consistency and developing a more efficient Gaussian initialization and optimization architecture remains an open area of research.
} 

{Several approaches have focused on improving Gaussian field consistency, particularly in sparse-view conditions.} CoR-GS \cite{zhang2024cor} introduced a co-pruning strategy to address inconsistencies in Gaussian representations that cannot be resolved through standard optimization methods. This approach establishes paired Gaussian fields to identify Gaussians that are inaccurately positioned. However, training two parallel fields significantly doubles the training time. {CoherentGS \cite{paliwal2024coherentgs} uses optical flow and monodepth priors to create a consistent, dense initialization of structured Gaussians, and an implicit decoder for smooth reconstruction. However, this pixel-aligned representation and dense initialization are inefficient for both training and storage, especially with sparse input views. While generalizable methods such as PixelSplat \cite{charatan2024pixelsplat} and MVSplat \cite{chen2024mvsplat} learn geometry from aggregated multi-view features, their performance often struggles with out-of-domain samples. Therefore, there's a strong need for a more efficient and robust method that can effectively solve the initialization and optimization challenges posed by sparse-view inputs.}

In this paper, we propose novel strategies for initialization and optimization to improve multi-view consistency, ensuring high rendering speed, short training duration, and low memory consumption. {We first introduce a sparse initializer that uses a sparse matcher and a random filling strategy to efficiently create a sufficient scene representation without causing significant disturbances. During optimization, we implement a multi-view consistency-guided progressive pruning strategy. This process begins by using a visual foundation model to extract general visual features for each Gaussian across all input views. We simplify the Gaussians to their central points, a technique that has proven effective for both efficient feature extraction and precise inconsistency detection.}
During the initial training stage, we compute a prune mask using high-level general visual features and gradually integrate low-level features in subsequent iterations. Our progressive pruning strategy targets Gaussians where the fine-grained color consistency constraint is weak and cannot be optimized or pruned effectively. This approach shifts the training focus toward the geometric entities while mitigating perturbations caused by randomly filled Gaussians in empty space, thereby reducing ``floater" artifacts common in sparse-view scenarios.
Together, the sparse initializer and pruning strategy enhance geometric consistency, contributing to both memory and rendering efficiency.
Extensive experiments with the LLFF, Blender, DTU, and MipNeRF360 datasets reveal that our method significantly improves performance over the vanilla 3DGS architecture in sparse-view scenarios, with fewer Gaussian primitives and higher rendering speed, making it a practical pipeline for sparse-view 3D Gaussian radiance fields.

Our main contributions are summarized as follows:

\begin{itemize}

    % \item We propose a novel sparse initializer that produces sparse yet sufficient initial points, which efficiently embeds geometric prior in the initialization process.

  \item {We propose a novel sparse initializer that generates a sufficient yet compact set of initial points. This approach enhances the efficiency of Gaussian representations while incorporating a robust geometric prior.}

    \item {We propose a multi-view consistency-guided progressive pruning strategy to shift the training focus to the geometric entity, leading to enhanced overall consistency.}

    \item {We explore the potential of a visual foundation model to enhance Gaussian consistency and develop an efficient pipeline for consistency checking using a simplified Gaussian representation.}

    \item The proposed MCGS serves as a plug-and-play module for 3DGS-based architectures that enhances their robustness to view sparsity, while also improving memory and rendering efficiency.

\end{itemize}

\section{Related Work}
\boldp{Radiance Fields}
Neural Radiance Fields (NeRFs) \cite{mildenhall2021nerf} have emerged as outstanding 3D representation methods for novel view synthesis. NeRF employs a large Multi-Layer Perceptron (MLP) network to map 3D coordinates to object attributes, such as radiance and color. Coupled with volume rendering, it produces remarkably photo-realistic novel views. Recent research has focused on improving rendering quality \cite{barron2021mip, barron2022mip}, enhancing efficiency \cite{muller2022instant}, and increasing robustness to sparse views \cite{yang2023freenerf}.
However, NeRF’s ray-marching process requires querying the MLP network for numerous sampling points, limiting its rendering speed. A recent alternative, 3D Gaussian Splatting (3DGS) \cite{kerbl20233d}, represents radiance fields using explicit Gaussian primitives and renders scenes through differentiable rasterization. This approach offers high-quality reconstruction, efficient training, and real-time rendering, making it well-suited for practical applications.
Despite these advantages, 3DGS relies on dense input views. Sparse input views pose challenges, as they result in insufficient Gaussian initialization from structure-from-motion (SFM) and reduce multi-view consistency, leading to geometry distortion and degraded novel view quality. Improving 3DGS performance under sparse-view conditions remains an open research problem.

\boldp{Novel View Synthesis Using Sparse Views}
Many methods have been proposed for NeRF to achieve realistic novel view synthesis from sparse input views. Some methods integrate additional 3D supervision, such as sparse point clouds \cite{deng2022depth} or estimated depth maps \cite{wang2023sparsenerf, roessle2022dense}, to regularize the geometry. In addition to utilizing extra 3D information, some approaches apply random patch-based semantic consistency constraints \cite{jain2021putting, seo2023mixnerf} or geometry regularization \cite{niemeyer2022regnerf}. Depth distillation \cite{wang2023sparsenerf} is a skillful technique for extracting depth priors from a depth estimation network, though it suffers from smoothness inherited from imprecisely estimated depth. GeCoNeRF \cite{kwak2023geconerf} introduces multi-view consistency regularization for few-shot novel view synthesis. Generative methods \cite{yu2021pixelnerf, chen2021mvsnerf, fei2024pixelgaussian, liu2024mvsgaussian, chen2024mvsplat, charatan2024pixelsplat} learn geometric priors from large multi-view datasets. These generalizable approaches aim to learn common patterns in scene geometry. While such methods have achieved strong performance on various multi-view datasets (e.g., DTU \cite{jensen2014large} or indoor datasets like RealEstate10K \cite{zhou2018stereo}), their generalization to complex real-world scenes—such as LLFF \cite{mildenhall2019local} or MipNeRF360 \cite{barron2022mip}—remains limited. When applied to new data domains, they typically require costly adaptation to maintain performance. In this work, we focus on developing a robust approach to address underfitted geometry in sparse-view settings—without relying on costly adaptation.

Another set of methods \cite{lao2024corresnerf, truong2023sparf} performs geometry regularization on estimated correspondences extracted from dense matchers. However, the training efficiency and rendering speed of these methods are still limited by their slow backbones. Although recent methods have shifted focus to designing strategies for grid-based backbones \cite{sun2023vgos}, they still face challenges with low rendering speed and a trade-off between training efficiency and rendering quality.

Recent methods \cite{li2024dngaussian, paliwal2024coherentgs, zhang2024cor, zhu2024fsgs} have been developed to enhance the performance of 3DGS when using sparse view input. FSGS \cite{zhu2024fsgs} leverages depth distillation to refine the geometry of a densely initialized Gaussian field derived from MVS, achieving notable improvements in sparse-view scenarios. 
{CoherentGS \cite{paliwal2024coherentgs} introduces a structured Gaussian representation that enables regularization directly in the 2D image space.} CoR-GS \cite{zhang2024cor} proposes a co-pruning strategy to remove inconsistent Gaussians and a co-regularization mechanism to correct geometric errors, establishing a new paradigm for pruning and refinement in sparse-view 3DGS. {Distinct from previous methods, our approach is the first to leverage robust general visual features from a large visual foundation model to enhance multi-view consistency. Instead of relying on inefficient dense initialization or time-consuming parallel training, our method improves robustness to view sparsity while achieving fast training and efficient scene representation.}

\section{Method}
\subsection{Preliminaries: 3D Gaussian Splatting}

\begin{figure*}[ht]

    \centering
    \includegraphics[width=1\linewidth]{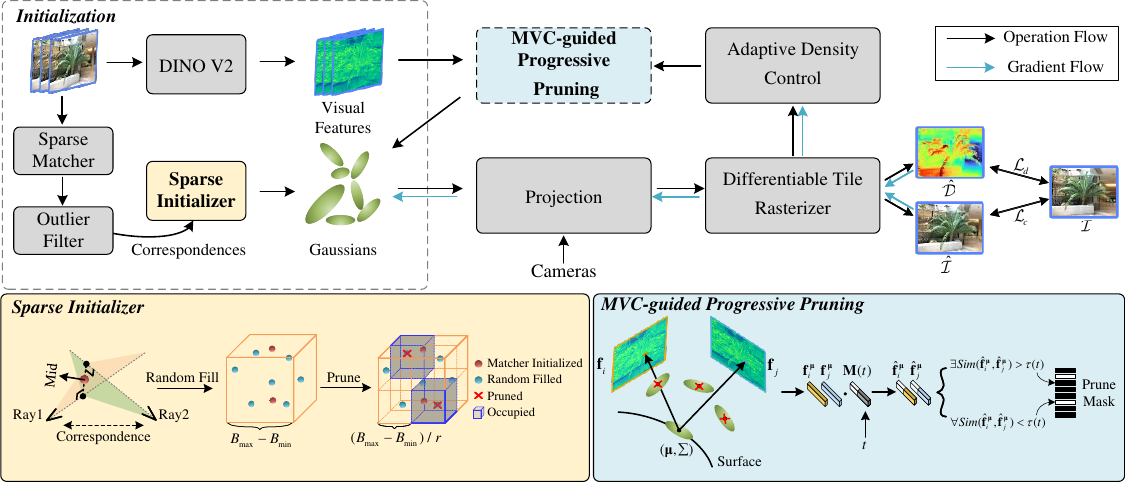}
    \caption{\textbf{An Overview of The Complete Framework.} Our proposed pipeline includes a sparse initializer to provide a sparse yet sufficient initial point cloud and a multi-view consistency-guided progressive pruning strategy to exclude noisy points from the initializer and enhance the overall multi-view consistency during training.}
    \label{fig:method}

\end{figure*}

\boldp{Representation and Rendering}
3D Gaussian Splatting (3DGS) \cite{shue20233d} represents a 3D scene as a set of anisotropic Gaussian primitives. Each primitive is defined by differentiable parameters: a position vector $\boldsymbol{\mu} \in \mathbb{R}^3$, a rotation quaternion $\mathbf{q} \in \mathbb{R}^4$, and a scaling vector $\mathbf{s} \in \mathbb{R}^3$. These parameters collectively define the primitive’s distribution in 3D space:
\begin{equation}
    G(\mathbf{x}) = e^{-\frac{1}{2}(\mathbf{x}-\boldsymbol{\mu})^T\boldsymbol{\Sigma}^{-1}(\mathbf{x}-\boldsymbol{\mu})},
\end{equation}
where the covariance matrix $\boldsymbol{\Sigma} \in \mathbb{R}^{3\times3}$ can be computed from the scale vector $\mathbf{s}$ and the rotation quaternion $\mathbf{q}$. These Gaussians are then rendered with their differentiable attributes, such as opacity $\alpha \in \mathbb{R}$ and color, which is represented by spherical harmonic coefficients.

3DGS utilizes the volume rendering function to render the Gaussians in order of their depth as
\begin{equation}
    \mathbf{c}(\mathbf{p}) = \sum_{i\in N}\mathbf{c}_i\hat{\alpha}_i \prod_{j=1}^{i-1}(1-\hat{\alpha}_j),
\end{equation}
where $\mathbf{p}$ is the overlapped pixel, $\mathbf{c}_i$ is the color of the $i$-th Gaussian $G_i(\mathbf{x})$ computed from spherical harmonic coefficients, and $\hat{\alpha}_i$ can be evaluated from the projected 2D Gaussian $G_i^{2D}(\mathbf{p})$:
\begin{equation}
    \hat{\alpha}_i = \alpha_i G^{2D}_i(\mathbf{p}).
\end{equation}
The pixel-wise depth $d$ can also be computed using the volume rendering function:
\begin{equation}
    d(\mathbf{p}) = \sum_{i\in N}d_i\hat{\alpha}_i\prod_{j=1}^{i-1}(1-\hat{\alpha}_j),
\end{equation}
where {$d_i = ||\boldsymbol{\mu}_i - \mathbf{o}||_2$} is the depth of the $i$-th Gaussian to the camera center $\mathbf{o}$.

\subsection{Sparse Initializer}
\label{sec:sparse_initial}
Current 3DGS techniques{\cite{kerbl20233d}} typically use point clouds derived from COLMAP for initializing 3D Gaussian representations. However, the point clouds generated by COLMAP are often sparse, leading to suboptimal performance, especially in scenarios with limited input views. To address this, some approaches incorporate multi-view stereo (MVS) methods to produce a denser point cloud during initialization{\cite{zhu2024fsgs, zhang2024cor}}. Despite this, the initial Gaussian primitives tend to exhibit high-frequency characteristics, introducing high-frequency artifacts \cite{yang2023freenerf} and resulting in inefficient scene representation.

In contrast, we utilize the end-to-end sparse matching network LightGlue \cite{lindenberger2023lightglue}, pre-trained on large-scale multi-view datasets, to obtain initial correspondences ${\langle \mathbf{r}_s, \mathbf{r}_t \rangle }_N$ for each image $\mathcal{I}_s$ and its paired view $\mathcal{I}_t$, where $\mathbf{r}_i = (\mathbf{o}_i, \mathbf{d}_i)$, $i \in \{s, t\}$, represents the ray with origin $\mathbf{o}_i$ and direction $\mathbf{d}_i$. {We employ a widely-adopted triangulation method to reconstruct a sparse point cloud from 2D correspondences, following the approach similar to CorresNeRF \cite{lao2024corresnerf}.} We compute the midpoint $\mathbf{p}_{mid}$ and its color $\mathbf{c}_{mid}$ for each correspondence pair $\langle \mathbf{r}_s, \mathbf{r}_t \rangle$ as:  $\mathbf{p}_{mid} = \frac{\mathbf{x}_s + \mathbf{x}_t}{2}$ and $\mathbf{c}_{mid} = \frac{\mathbf{c}_s + \mathbf{c}_t}{2}$, where $\mathbf{x}_s$ and $\mathbf{x}_t$ are the nearest points on the two rays, and $\mathbf{c}_s$ and $\mathbf{c}_t$ are their respective colors. This results in an initialized point cloud ${\langle \mathbf{p}_{mid}, \mathbf{c}_{mid} \rangle }_N$. Optionally, for datasets with significant symmetry that may cause visual ambiguity, we apply an outlier filter following CorresNeRF \cite{lao2024corresnerf}.

\boldp{Random Filling}
The sparse feature matching process encounters substantial challenges in low-texture areas or regions with repetitive structures \cite{lindenberger2023lightglue, sarlin2020superglue}, often resulting in insufficiently initialized Gaussians. {CorresNeRF \cite{lao2024corresnerf} enhances correspondence quality through image augmentation and subsequently applies an interpolation-based expansion strategy to extend the spatial support of each correspondence. However, it still struggles to generate sufficient initial points in low-texture regions.} To mitigate this issue, we propose a random filling strategy. Initially, we randomly initialize $N$ points within the bounding box defined by $(B_{min}, B_{max})$. To prevent interference with points initialized by the sparse matching process, we exclude any randomly generated points that fall within voxels already occupied by the points derived from the sparse matcher. The voxel size is calculated as $\frac{B_{max} - B_{min}}{r}$, where $r$ represents the predefined resolution. This random filling approach enhances the overall completeness of the initialized geometry while minimizing disruption to points containing multi-view consistency priors. {Following the methodology proposed in 3DGS \cite{kerbl20233d}, we transform the fully initialized point cloud—including both matcher-initialized and randomly filled points—into a set of initial 3D Gaussians.} However, some randomly filled points may reside in empty spaces with minimal contribution to geometric reconstruction, which can result in suboptimal artifacts. These extraneous points are subsequently addressed through a newly designed pruning strategy.

\subsection{Multi-view Consistency Guided Progressive Pruning}
Current 3DGS methods{\cite{kerbl20233d, yu2024mip}} often utilize multi-view consistency constraints based on color to optimize and densify Gaussians. Such fine-grained constraints struggle with inconsistent Gaussians, particularly in low-texture regions where spatial similarity in color leads to ambiguity, especially when input views are sparse. Recently, general features extracted from DINOv2 \cite{oquab2023dinov2}, embedded with global context, have demonstrated robust performance in identifying slightly inconsistent structures. Despite this, integrating these visual features effectively into the 3DGS to handle inconsistent Gaussians remains a challenge. {The recent NeRF-based method, NeRF-On-the-Go \cite{ren2024nerf}, leverages priors from DINO to distinguish transient object features across different views. However, it relies on strong contrastive loss patterns between dynamic distractors and the static background in the training views. In sparse-view settings, such high-contrast patterns rarely emerge due to overfitting to the limited training data.}

To overcome these challenges, we propose a multi-view consistency-guided (MVC-guided) progressive pruning strategy, utilizing the powerful DINOv2 visual encoder to exclude inconsistent Gaussians. We begin by extracting visual features $\{\mathbf{f}_i, i=1,...,N\}$ from $N$ input images $\{\mathcal{I}_i, i=1,...,N\}$. Each feature $\mathbf{f}_i \in \mathbb{R}^{K \times H \times W}$ contains $K$-dimensional representations, concatenated from components of dimensions $\{K_l, l=1,...,L\}$ extracted from various levels $l$ of DINOv2. {We upsample the features to the pixel scale and then extract per-view features using projected Gaussians. To improve the efficiency and precision of this process, we simplify the Gaussians to their central points. This simplification is crucial for our multi-view consistency assessment, and we provide a detailed analysis of its impact in Sec. \ref{sec:analysis}.} For Gaussians with central positions $\{\boldsymbol{\mu}_j, j=1,...,M\}$, we project the 3D points onto the input images using intrinsic and extrinsic parameters $\mathbf{K}_i$ and $(\mathbf{R}_i, \mathbf{T}_i)$:
\begin{equation}
    \label{eq:uproject}
    {\boldsymbol{\mu}}^j_i = \frac{1}{z}\mathbf{K}_i(\mathbf{R}_i\boldsymbol{\mu}_j +\mathbf{T}_i),
\end{equation}
where $z$ is the distance to the camera origin along the $z$ axis in the camera coordinate system. We query the feature $\mathbf{f}^j_i \in \mathbb{R}^{K}$ at position ${\boldsymbol{\mu}}^j_i$ from the feature map $\mathbf{f}_i$ using bilinear interpolation. For each Gaussian, we compute the cosine similarity $s$ between features $\langle \mathbf{f}_m^{j}, \mathbf{f}_n^{j} \rangle$ queried from pairs of images, with $m \neq n$, as $s_{mn}^{j} = \frac{\mathbf{f}_m^{j} \cdot \mathbf{f}_n^{j}}{||\mathbf{f}_m^{j}||_2 ||\mathbf{f}_n^{j}||_2}$. We apply the mask $\mathbf{M}$ to prune Gaussians as
\begin{equation}
    M_j = 
    \begin{cases} 
        1, & \text{if } s_{mn}^{j} < \tau,  \forall m, n \in [N], \ m \neq n \\
        0, & \text{otherwise}
    \end{cases},
    \label{eq:mask1}
\end{equation}
where $[N]$ represents the set of integers from $1$ to the number of input views $N$, and $M_j$ is the $j$-th element in $\mathbf{M}$ and the $j$-th Gaussian. Directly applying fixed levels of features and a threshold $\tau$ leads to insufficient Gaussians during the initial training phase. As a result, we relax the constraint and design a progressive pruning strategy by applying masks to the features to exclude low-level features and gradually incorporate these low-level features during training. Additionally, we increase the threshold $\tau$ from a small initial value. The mask represented by Eq. \ref{eq:mask1} can be adapted as
\begin{equation}
    M_j = 
    \begin{cases} 
        1, & \text{if } s_{mn}^{j}(t) < \tau(t), \forall m, n \in [N], \ m \neq n \\
        0, & \text{otherwise}
    \end{cases}.
    \label{eq:mask2}
\end{equation}
We apply the mask $\hat{\mathbf{M}}(t)$ to the features to compute similarity with a progressively increased pruning step $t \in [T]$:
\begin{equation}
    \label{eq:sim_compute}
    s_{mn}^{j}(t) = \frac{(\hat{\mathbf{M}}(t)\mathbf{f}_m^{j}) \cdot  (\hat{\mathbf{M}}(t)\mathbf{f}_n^{j})}{||\hat{\mathbf{M}}(t)\mathbf{f}_m^{j}||_2||\hat{\mathbf{M}}(t)\mathbf{f}_n^{j}||_2}
\end{equation}
with
\begin{equation}
    \hat{{M}}_k(t) = 
    \begin{cases} 
        1, & k>\sum_{l=1}^{L-t}K_{l},t<L \\
        1, & t \geq L \\
        0, &  \text{otherwise}
    \end{cases},
\end{equation}
where $\hat{M}_k(t)$ is the $k$-th element in the mask $\hat{\mathbf{M}}(t)$, and $\tau(t)$ is set as a hyperparameter. The progressive pruning strategy excludes Gaussians in empty space that have low contributions to the rendered results, thereby focusing training and densification more on Gaussians on the surface.

\subsection{Edge-Aware Depth Regularization}
\begin{figure}[t]

    \centering
    \includegraphics[width=1\linewidth]{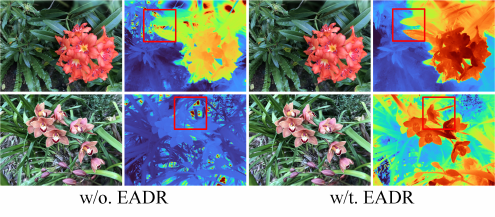}
    \caption{We present the RGB and depth results generated by our method, comparing scenarios with and without edge-aware depth regularization (EADR). Geometric voids that cannot be resolved by our pruning strategy are effectively addressed using the EADR approach.}
    \label{fig:progressive}

\end{figure}

The proposed method improves the multi-view consistency of Gaussian representations by utilizing general visual features. However, the newly introduced pruning strategy encounters challenges with geometric voids (see Fig. \ref{fig:progressive}). These voids arise from either an insufficient number of initial Gaussians or pruning errors based on opacity, especially in low-texture regions.
To address these issues, we build on the techniques presented in \cite{heise2013pm,comi2024snap}, incorporating edge-aware depth regularization (EADR) to correct geometric distortions. After the final MVC-guided progressive pruning step $T$, we apply EADR to enhance depth continuity and minimize artifacts along object boundaries.
\begin{equation}
    \begin{aligned}\mathcal{L}_d=\frac{1}{N}\omega_d\sum_{i,j}\left(|\partial_xD_{i,j}|e^{-\beta|\partial_x\mathcal{I}_{i,j}|}+|\partial_yD_{i,j}|e^{-\beta|\partial_y\mathcal{I}_{i,j}|}\right)\end{aligned},
\end{equation}
where $D_{i,j}$ represents the $(i,j)$-th pixel of the depth map $\mathbf{D}$, and $\mathcal{I}_{i,j}$ denotes the corresponding pixel in the input image. The parameter $N$ is the total number of pixels, and $\beta$—fixed at 2 throughout all experiments—acts as a hyperparameter controlling edge sharpness in the depth map. During the initial training phase, the weight $\omega_d$ is set to 0. After the final pruning step, it is increased to 1.
\begin{equation}
    \omega_d = 
    \begin{cases} 
        0, & i<(T-1)i_{step} \\
        % 1, & t \geq L \\
        1, &  \text{otherwise}
    \end{cases},
\end{equation}
where $i$ is the current iteration count, $T$ denotes the total number of pruning steps, and $i_{step}$ specifies the number of iterations between consecutive pruning steps. The refinement process aims to minimize geometry errors in low-texture regions while smoothing the previously established geometry.

\begin{figure*}[ht]

    \centering
    \includegraphics[width=1\linewidth]{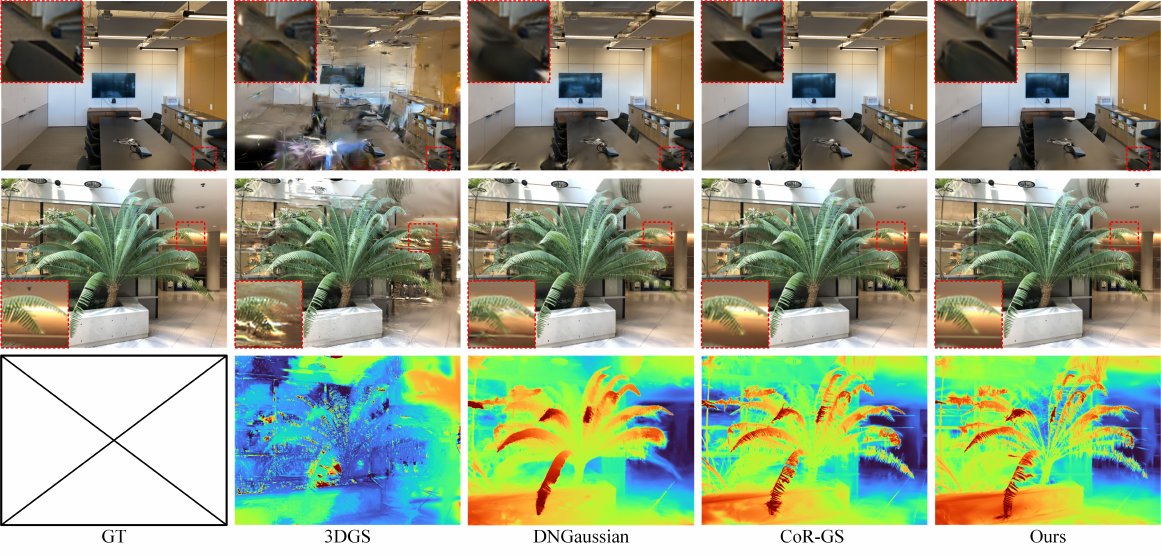}
    \caption{\textbf{Qualitative Comparison on LLFF.} We present novel views of the scenes ``room" and ``fern" (first and second rows) rendered by our method as well as the baselines: 3DGS, DNGaussian, and CoR-GS. Our method achieves the most realistic novel views, especially in the areas highlighted by red boxes. Please enlarge the PDF to see more details.}
    \label{fig:llff}

\end{figure*}

\section{Experiments}
\subsection{Setups}
\boldp{Datasets and Metrics}
We evaluate our method on four benchmark datasets: the synthetic Blender dataset \cite{mildenhall2021nerf}, the forward-facing LLFF dataset \cite{mildenhall2019local}, the DTU dataset \cite{jensen2014large}, and the MipNeRF360 dataset \cite{barron2022mip}.
For the Blender dataset, we follow the protocols established by FreeNeRF \cite{yang2023freenerf} and DietNeRF \cite{jain2021putting}. We train our model on a small subset of views (e.g., 4 or 8 views, with specific IDs: 26, 86, 2, 55, 75, 93, 16, and 73) and evaluate its performance on 25 test images that are uniformly sampled from the test set, all downsampled by a factor of 2.
On the LLFF dataset, we adhere to the procedure outlined in FreeNeRF. We evaluate on every eighth image and train our model using the first $N$ images (e.g., 3 or 6) from the remaining views. All images in this experiment are downsampled by a factor of 8.
For the MipNeRF360 dataset, we select either 12 or 24 training views by following the same training split procedure used for the LLFF dataset.
Finally, for the DTU dataset, we again follow the experimental setup of FreeNeRF. We conduct experiments with 3 and 6 input views and apply an 8× downsampling rate. Evaluations are performed on images masked with ground truth object masks to focus on the object of interest.
To ensure a fair comparison and align with community standards, we report PSNR, SSIM, and LPIPS metrics, with results averaged across all scenes.

{To comprehensively evaluate our method against alternative generalizable approaches (e.g., MVSplat \cite{chen2024mvsplat}), we implemented our model on three additional complex, real-world datasets: RealEstate10K (RE10K) \cite{zhou2018stereo}, ACID \cite{liu2021infinite}, and DL3DV \cite{ling2024dl3dv}. We conducted these experiments at a resolution of 256x256, following MVSplat's original settings.}

\boldp{Implementation Details}
We implement MCGS using the 3DGS base architecture \cite{kerbl20233d}. The initialized point cloud is obtained through our sparse initializer, which relies solely on the training views. We utilize calibrated ground truth poses in the datasets, adhering to community standards. We set the total number of iterations to 10,000, with the position learning rate decreasing from 0.0016 to 0.000016. The densification interval and the opacity reset interval are set to 300 and 1,000, respectively. The density gradient threshold is fixed at 0.0005. For our proposed method, we configure the total pruning steps $T$ to 4, with the pruning step increasing by 1 every 3,000 iterations for the LLFF or Blender dataset, respectively. For feature similarity computation, we extract features from DINO-ResNet-50 \cite{oquab2023dinov2} with feature dimensions $K_l \in \{64, 64, 128, 256\}$. The pruning threshold $\tau(t)$ is represented by the lists $[0.7, 0.75, 0.8, 0.85]$ and $[0.6, 0.65, 0.7, 0.8]$ for the LLFF and Blender datasets, respectively. For the Blender dataset, we apply an outlier filter to remove inconsistent correspondences according to CorresNeRF \cite{lao2024corresnerf} and use background occlusion regularization following FreeNeRF \cite{yang2023freenerf}. All experiments are conducted on a single NVIDIA 3090 GPU. 

\subsection{Evaluation on the LLFF Dataset}
We present comprehensive qualitative and quantitative evaluations on the LLFF dataset, as illustrated in Fig. \ref{fig:llff} and Tab. \ref{tab:llff}. Qualitatively, our method significantly outperforms the 3DGS baseline \cite{kerbl20233d}. While the base 3DGS architecture suffers from noticeable distortion, our approach produces highly realistic renderings with fine-grained details. Compared to DNGaussian \cite{li2024dngaussian}—a few-shot novel view synthesis (NVS) baseline derived from 3DGS—our method avoids the blurred outputs caused by its coarse-grained depth regularization (which relies on monocular depth estimation). Crucially, we preserve high-frequency details without dependence on pre-trained depth networks. Even against the recent CoR-GS \cite{zhang2024cor}, our renderings exhibit superior accuracy and visual fidelity.

\begin{table}[!htp]

\centering

    \caption{\textbf{Quantitative Comparison on LLFF.} The best and second-best quantitative results are marked in red and grey, respectively.}

    \setlength{\tabcolsep}{3pt}
    \begin{tabular}{l|ccc|ccc}  \hline
    \multirow{2}{*}{Method}  & \multicolumn{3}{c|}{3 Views} & \multicolumn{3}{c}{6 Views}\\ & PSNR$\uparrow$ & SSIM$\uparrow$ & LPIPS$\downarrow$ & PSNR$\uparrow$ & SSIM$\uparrow$ & LPIPS$\downarrow$ \\ \hline

    RegNeRF \cite{niemeyer2022regnerf}    & {19.08}         & {0.587}          & {0.336}     & 23.10 & 0.760 & 0.206  \\ 
    
    FreeNeRF \cite{yang2023freenerf}  &  {19.63}          &  {0.612}          & 0.308 & 23.73 & 0.779 & 0.195 \\

    SparseNeRF \cite{wang2023sparsenerf}      &  {19.86}          &  {0.624}          &  0.328 & - & - & - \\

    SPARF \cite{truong2023sparf}     &  {20.20}          &  {0.630}          &  0.327 & 23.35 & 0.740 & 0.200 \\ \hline
    % 3DGS \cite{kerbl20233d}      &  {14.94} &  {0.468} & {0.379}  \\
    3DGS \cite{kerbl20233d}      &  {19.09} &  {0.645} & {0.242}  & 23.89 & {0.815} & {0.148}\\

    DNGaussian  \cite{li2024dngaussian}    &  {20.02} &  {0.682} & {0.229} & 22.83 & 0.777 & 0.182  \\ 
    FSGS  \cite{zhu2024fsgs}    &  \cellcolor{red!25}{20.37} &  \cellcolor{grey!25}{0.696} & \cellcolor{red!25}{0.205} & \cellcolor{grey!25}{24.10} & \cellcolor{grey!25}{0.818} &  \cellcolor{red!25}{0.127}  \\ \hline

    Ours     &  \cellcolor{grey!25}{20.33} &  \cellcolor{red!25}{0.699} & \cellcolor{grey!25}{0.226}    & \cellcolor{red!25}{24.28} & \cellcolor{red!25}{0.823} & \cellcolor{grey!25}{0.139}   \\ \hline

    \end{tabular}

\label{tab:llff}

\end{table}

The quantitative results further verify the effectiveness of our method. Our method surpasses the base 3DGS by 1.2 dB with our setting. Both qualitative and quantitative results indicate that our method significantly enhances the multi-view consistency of 3D Gaussians. 

% We also integrate our method into the recent MipSplatting architecture \cite{yu2024mip} to verify its effectiveness across different architectures. Our method achieves a 1.2 dB PSNR improvement compared to MipSplatting with our setting.

\subsection{Evaluation on the Blender Dataset}
\begin{figure}[!htp]

    \centering
   \includegraphics[width=1\linewidth]{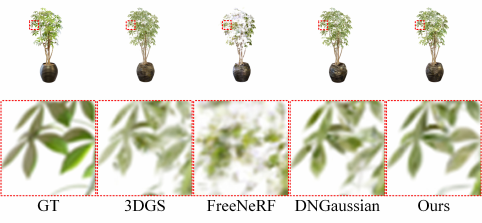}
    \caption{\textbf{Qualitative Comparison on Blender Using 8 Input Views.} We present novel views and zoomed-in areas (highlighted by red boxes) rendered by our method and the baselines.}
    \label{fig:blender1}

\end{figure}

Fig. \ref{fig:blender1} and Tab. \ref{tab:blender1} show the qualitative and quantitative results, respectively. For qualitative results, we compare our method with the base 3DGS \cite{kerbl20233d}, the well-known baseline FreeNeRF \cite{yang2023freenerf}, and DNGaussian \cite{li2024dngaussian}. Our method shows comparable results to FreeNeRF and DNGaussian, while performing better in areas highlighted by red boxes. {For quantitative results, our method surpasses the base architecture 3DGS by 1.3 dB and DNGaussian by 0.1 dB in PSNR when using 8 input views. For extremely sparse settings with only 4 input views, our method achieves an even more significant improvement of over 2 dB.} These qualitative and quantitative results further verify the effectiveness of our method on the panoramic dataset.

\begin{table}[!htp]

    \centering

    \caption{\textbf{Quantitative Comparison on Blender.} The best and second-best quantitative results are marked in red and grey, respectively.}

    \setlength{\tabcolsep}{3pt}
    \begin{tabular}{l|ccc|ccc}
    
    \hline
    \multirow{2}{*}{Method}  & \multicolumn{3}{c|}{8 Views} & \multicolumn{3}{c}{4 Views}\\ & PSNR$\uparrow$ & SSIM$\uparrow$ & LPIPS$\downarrow$ & PSNR$\uparrow$ & SSIM$\uparrow$ & LPIPS$\downarrow$ \\ 
    \hline

    DietNeRF \cite{jain2021putting}  &  23.15 & 0.866 & 0.109 &  15.42 & 0.730 & 0.314\\

    InfoNeRF \cite{kim2022infonerf}&  22.01 & 0.852 & 0.133 &  18.44 & 0.792 & 0.223 \\
    MixNeRF \cite{seo2023mixnerf} &  23.84 & 0.878 & 0.103 &  \cellcolor{grey!25}{18.99} & \cellcolor{grey!25}{0.807} & 0.199  \\

    FreeNeRF\cite{yang2023freenerf} &  \cellcolor{red!25}{24.26} & \cellcolor{grey!25}{0.883} &   \cellcolor{grey!25}{0.098}   &  - & - & - \\

    \hline
    3DGS \cite{kerbl20233d} & {22.77} & {0.865} & 0.112 &  17.68 & 0.772 & 0.201 \\
    DNGaussian\cite{li2024dngaussian} & {23.90} & {0.880} & \cellcolor{red!25}{0.089} &  17.78  & 0.770 & \cellcolor{grey!25}{0.186} \\
    Ours  &  \cellcolor{grey!25}{24.06} & \cellcolor{red!25}{0.887} & \cellcolor{red!25}{0.089} &  \cellcolor{red!25}{19.85} & \cellcolor{red!25}{0.823} & \cellcolor{red!25}{0.147} \\

    \hline
    \end{tabular}

    \label{tab:blender1}

\end{table}

\subsection{Evaluation on the DTU Dataset}
\begin{figure}[!htp]

    \centering
    \includegraphics[width=1\linewidth]{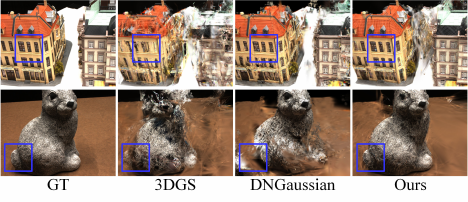}

    \caption{\textbf{Qualitative Comparison on DTU.} We show novel views rendered by our method and baseline methods in a 3-view setting. The detailed results, highlighted in blue boxes, demonstrate significant improvement.}
    \label{fig:DTU}

\end{figure}

Fig. \ref{fig:DTU} and Tab. \ref{tab:DTU} present the qualitative and quantitative results on the DTU dataset, respectively. For quantitative results, our method shows significant improvement compared to the base 3DGS architecture \cite{kerbl20233d} (surpassing 3DGS by 3 dB in terms of PSNR). Moreover, {our method outperforms the recent 3DGS-based few-shot novel view synthesis method, DNGaussian \cite{li2024dngaussian}, in a 3-view setting. It achieves a more significant improvement of nearly 1.5 dB over DNGaussian in a 6-view setting.} The qualitative results further confirm the improvement, with our method producing more detailed rendered results with less distortion, as highlighted in the blue box in Fig. \ref{fig:DTU}.

\begin{table}[!htp]
    \centering
    \caption{\textbf{Quantitative Comparison on DTU.} The best and second-best quantitative results are marked in red and grey, respectively.}

    \setlength{\tabcolsep}{3pt}
    \begin{tabular}{l|ccc|ccc}  \hline
    \multirow{2}{*}{Method}  & \multicolumn{3}{c|}{3 Views} & \multicolumn{3}{c}{6 Views}\\ & PSNR$\uparrow$ & SSIM$\uparrow$ & LPIPS$\downarrow$ & PSNR$\uparrow$ & SSIM$\uparrow$ & LPIPS$\downarrow$ \\ 
    \hline

    PixelNeRF ft \cite{yu2021pixelnerf} &  {18.95} & {0.710} & {0.269} & 20.56  &  0.753 & 0.223\\
    MVSNeRF ft \cite{mvsnerf}&  {18.54} & {0.769} & {0.197}  &  20.49 &  0.822 &  0.155\\

    DietNeRF \cite{jain2021putting} &{11.85} & {0.633} & {0.314}  & 20.63  & 0.778 & 0.201\\
    RegNeRF \cite{niemeyer2022regnerf}  &  {18.89} & {0.745} & {0.190}   &  22.20 &  0.841 & \cellcolor{grey!25}{0.117}\\ 
    FreeNeRF   \cite{yang2023freenerf}        &  \cellcolor{red!25}{19.92} & {0.787} & {0.182} & \cellcolor{grey!25}{23.25}  & 0.844 &  0.137 \\\hline
    3DGS \cite{kerbl20233d} &  {14.92} & {0.698} & {0.248}  & 21.05  & 0.844 & 0.141 \\
    DNGaussian \cite{li2024dngaussian} &  {18.91} & \cellcolor{grey!25}{0.790} & \cellcolor{grey!25}{0.176}  & {22.93} & \cellcolor{grey!25}{0.872} & 0.129 \\
    Ours  &  \cellcolor{grey!25}{19.02} & \cellcolor{red!25}{0.810} & \cellcolor{red!25}{0.154}  & \cellcolor{red!25}{24.25}  & \cellcolor{red!25}{0.917} & \cellcolor{red!25}{0.084} \\

    \hline
    \end{tabular}

    \label{tab:DTU}

\end{table}

\subsection{Evaluation on the MipNeRF360 Dataset}
\begin{figure*}
    \centering
    \includegraphics[width=1\linewidth]{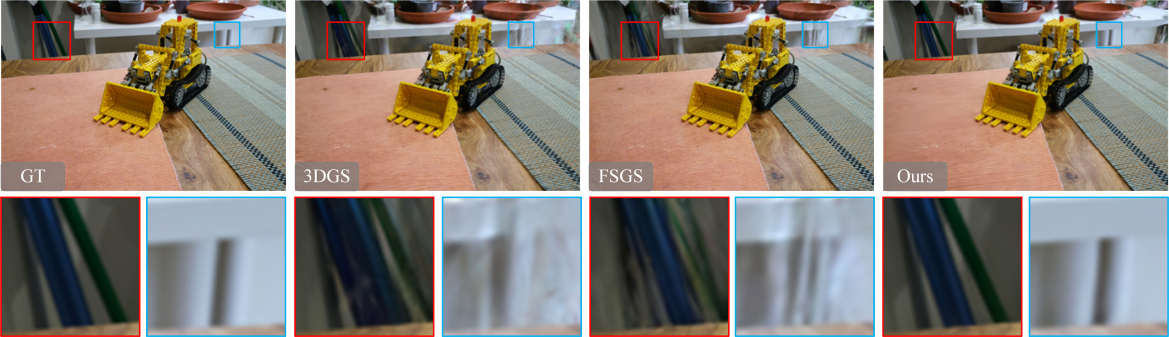}
    \caption{\textbf{Qualitative Comparison on MipNeRF360.} Our proposed method demonstrates superior rendering performance compared to the baseline approaches (3DGS and FSGS), as evidenced by the novel view synthesis results in large-scale real-world scenarios. The comparative visualization clearly validates the effectiveness of our approach.}
    \label{fig:mip360}
\end{figure*}

{To further assess the effectiveness of our approach in large-scale, real-world scenarios, we conducted additional experiments on the MipNeRF360 dataset \cite{barron2022mip}. The quantitative results are summarized in Tab. \ref{tab:360}, while the qualitative comparison is illustrated in Fig. \ref{fig:mip360}. These quantitative evaluations show that our method delivers substantial improvements over the original 3DGS baseline \cite{kerbl20233d}. Moreover, our performance is comparable to that of the recent FSGS method \cite{zhu2024fsgs}. As shown in Fig.~\ref{fig:mip360}, the qualitative results further demonstrate the advantages of our method, particularly in zoomed-in regions where finer details are more accurately preserved.}

\begin{table}[!htp]

    \centering

    \caption{\textbf{Quantitative Comparison on MipNeRF360.} The best and second-best quantitative results are marked in red and grey, respectively.}

    \setlength{\tabcolsep}{3pt}
    \begin{tabular}{l|ccc|ccc}
    
    \hline
    \multirow{2}{*}{Method} & \multicolumn{3}{c|}{12 views}  & \multicolumn{3}{c}{24 views}\\ 
    & PSNR$\uparrow$ & SSIM$\uparrow$ & LPIPS$\downarrow$ & PSNR$\uparrow$ & SSIM$\uparrow$ & LPIPS$\downarrow$ \\ \hline

    3DGS  \cite{kerbl20233d}  & 18.52 &  0.523 & {0.415} & 22.77 & 0.696 & 0.290   \\
    FSGS  \cite{zhu2024fsgs}       & \cellcolor{grey!25}{19.18} &  \cellcolor{grey!25}{0.552} & \cellcolor{red!25}{0.367}   & \cellcolor{grey!25}{23.38} &  \cellcolor{grey!25}{0.732} &  \cellcolor{red!25}{0.238} \\

    Ours  & \cellcolor{red!25}{19.36} & \cellcolor{red!25}{0.554} & \cellcolor{grey!25}{0.389} & \cellcolor{red!25}{23.39} & \cellcolor{red!25}{0.735} & \cellcolor{grey!25}{0.278}   \\

    \hline
    \end{tabular}

    \label{tab:360}

\end{table}

\subsection{Comparison with Generalizable Methods}
{Generalizable methods \cite{yu2021pixelnerf, chen2021mvsnerf, fei2024pixelgaussian, liu2024mvsgaussian, chen2024mvsplat, charatan2024pixelsplat} for 3D reconstruction and novel view synthesis, which rely on feed-forward pipelines, have shown impressive performance on in-domain datasets. These approaches learn common geometric and appearance patterns from extensive datasets. While recent methods like PixelSplat \cite{charatan2024pixelsplat} and MVSplat \cite{chen2024mvsplat} use cross-view-aware features to simplify geometry learning, their performance suffers significantly when applied to data with a substantial domain gap from the training data. This limitation restricts their use in real-world applications. To address this, we propose a more robust, regularization-based method designed for under-constrained, per-scene optimization of 3DGS with sparse input views. We demonstrate our method's stability and robustness by comparing it against the state-of-the-art generalizable method, MVSplat, across four diverse datasets:}
\begin{itemize}
    \item {\textbf{RealEstate10K (RE10K) and ACID}: We utilized small subsets of these in-domain datasets, which consisted of 38 and 13 scenes, respectively. These subsets were split by MVSplat, and we followed its protocols for selecting input and novel views.}

    \item {\textbf{DL3DV}: We performed experiments on a subset of 11 calibrated scenes, which were split by the official DL3DV. For training, we used the first and last frames from the initial 40 frames. Three test views were then uniformly sampled from the frames in between.}

    \item {\textbf{LLFF}: We selected one sample every four frames from the LLFF dataset. The start and end frames were designated as the training views, and the three intermediate frames were used for testing.}
\end{itemize}

\begin{figure*}
    \centering
    \includegraphics[width=1\linewidth]{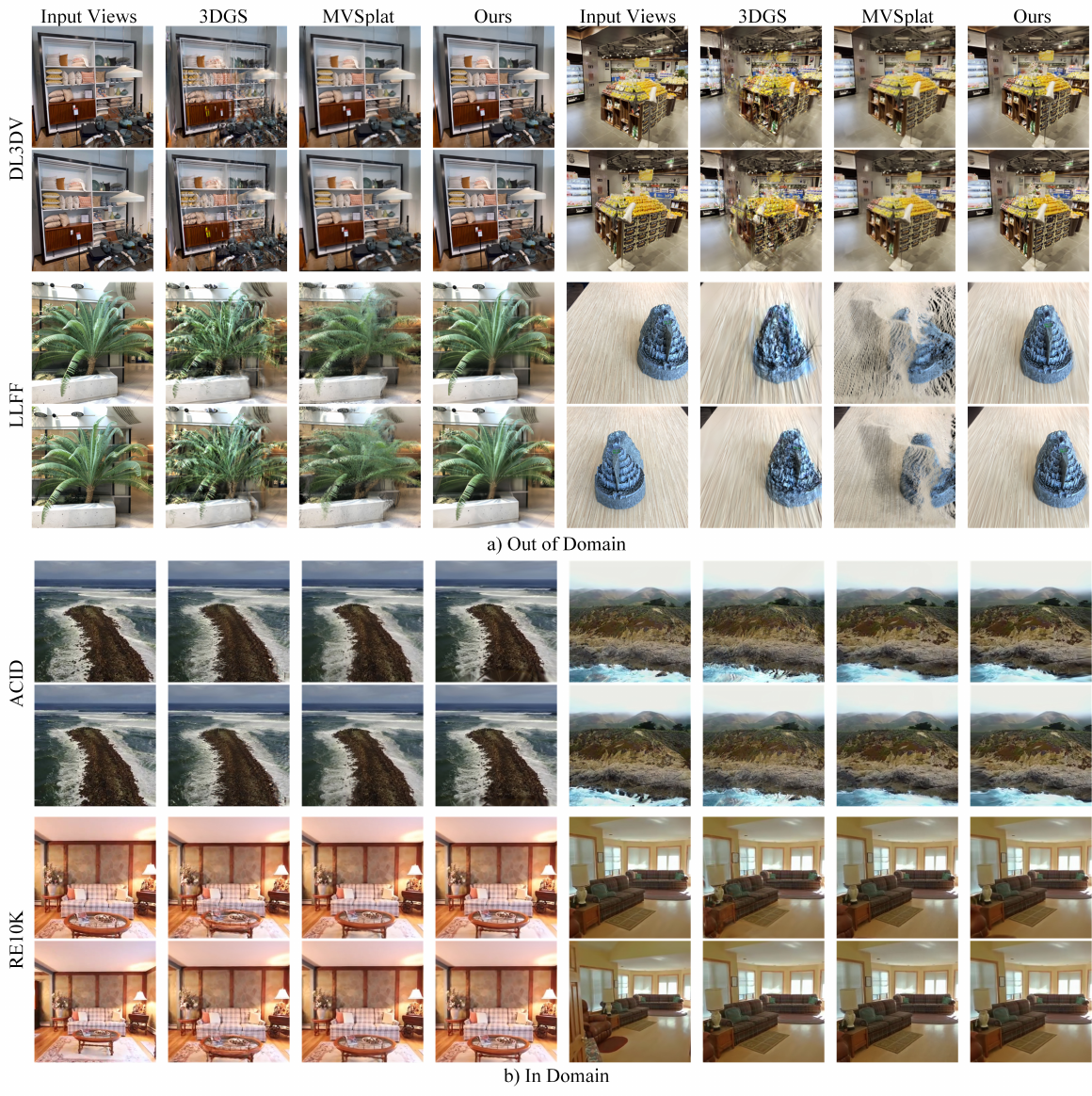}
    \caption{\textbf{Qualitative Comparison with Generalizable Methods.}}
    \label{fig:general}
\end{figure*}

{Tab. \ref{tab:general} and Fig. \ref{fig:general} present the quantitative and qualitative comparisons of our method with the generalizable method, MVSplat, and the 3DGS baseline. The quantitative results show that while MVSplat surpasses our method on its specifically trained in-domain datasets (RE10K and ACID), our approach yields more stable results on out-of-domain datasets, such as DL3DV and LLFF. Specifically, our method demonstrates a 2 dB PSNR improvement on the DL3DV dataset and a 4 dB improvement on the LLFF dataset over MVSplat. These findings are supported by the qualitative results, which underscore the stability and robustness of our method. Notably, the feed-forward MVSplat method introduces significant distortion and blur on the LLFF and DL3DV datasets, whereas our method produces significantly more accurate reconstructions and realistic renderings.}

\begin{table}[!htp]
    % \vskip -0.1in
    \centering
    \small
    \setlength{\tabcolsep}{3pt}
    \caption{\textbf{Quantitative Comparison with Generalizable Methods.} We conduct experiments on two in-domain datasets and two out-of-domain datasets.}

    \begin{tabular}{l|ccc|ccc}
    
    \hline
    \multicolumn{7}{c}{\textbf{In Domain}} \\ \hline
     & \multicolumn{3}{c|}{\textbf{RE10k}} &\multicolumn{3}{c}{\textbf{ACID}}\\ 
     & PSNR$\uparrow$ & SSIM$\uparrow$ & LPIPS$\downarrow$ & PSNR$\uparrow$ & SSIM$\uparrow$ & LPIPS$\downarrow$\\  \hline
     3DGS \cite{kerbl20233d} & 20.14 & 0.683 & 0.234 & 22.96 & 0.701 & 0.210\\ 
     MVSplat \cite{chen2024mvsplat} & \cellcolor{red!25}{26.02} & \cellcolor{red!25}{0.874} & \cellcolor{red!25}{0.119} & \cellcolor{red!25}{27.83} & \cellcolor{red!25}{0.871} & \cellcolor{red!25}{0.123}\\ 
     Ours & \cellcolor{grey!25}{23.64} & \cellcolor{grey!25}{0.829} & \cellcolor{grey!25}{0.170} & \cellcolor{grey!25}{25.78} & \cellcolor{grey!25}{0.834} & \cellcolor{grey!25}{0.185}\\ \hline

    \multicolumn{7}{c}{\textbf{Out of Domain}} \\ \hline
     & \multicolumn{3}{c|}{\textbf{DL3DV}} &\multicolumn{3}{c}{\textbf{LLFF}}\\ 
     & PSNR$\uparrow$ & SSIM$\uparrow$ & LPIPS$\downarrow$ & PSNR$\uparrow$ & SSIM$\uparrow$ & LPIPS$\downarrow$\\  \hline
     3DGS \cite{kerbl20233d} & 15.34 & 0.399 & {0.403} & 16.18 & 0.444 & \cellcolor{grey!25}{0.349}\\ 
     MVSplat \cite{chen2024mvsplat} & \cellcolor{grey!25}{18.03} & \cellcolor{grey!25}{0.499} & \cellcolor{grey!25}{0.365} & \cellcolor{grey!25}{16.70} & \cellcolor{grey!25}{0.462} & 0.391\\ 
     Ours & \cellcolor{red!25}{20.08} & \cellcolor{red!25}{0.667} & \cellcolor{red!25}{0.259} & \cellcolor{red!25}{20.73} & \cellcolor{red!25}{0.690} & \cellcolor{red!25}{0.251}\\ 

    \hline
    \end{tabular}

    \label{tab:general}

\end{table}

\subsection{Method Analysis \& Ablation Study}
\label{sec:analysis}
% \boldp{Efficiency Assessment}
\subsubsection{Efficiency Assessment}
In this study, we propose a novel sparse initializer and a pruning strategy to enhance multi-view consistency during initialization and optimization. Our method not only significantly improves the overall consistency of the field but also represents the scene more efficiently with fewer Gaussians, leading to higher memory efficiency and faster rendering speeds. Tab. \ref{tab:efficiency} presents the assessment of training time (Time), FPS, number of Gaussians (\#GS(k)), and visual quantitative results (PSNR, SSIM, LPIPS). The results indicate that our method achieves the best visual performance with far fewer Gaussians and offers faster training (7$\times$ faster than FSGS \cite{zhu2024fsgs}) and rendering speeds compared to the FSGS method. Notably, our method surpasses the base 3DGS architecture \cite{kerbl20233d}, implemented with our setting, by 1.2 dB in PSNR and achieves nearly 1.5$\times$ faster rendering speed with minimal training efficiency loss (less than 20 s), which is due to the additional pruning step.

\begin{table}[!htp]

    \centering

    \caption{\textbf{Efficiency Comparison.} Our method achieves efficient training, real-time rendering, and optimal results with a minimal number of Gaussian primitives. The best and second-best quantitative results are marked in red and grey, respectively.}
    \setlength{\tabcolsep}{1mm} 
    {
    \begin{tabular}{l|cccccc}
    
    \hline
    Method & Time$\downarrow$ & FPS$\uparrow$ & \#GS(k)  & PSNR$\uparrow$ & SSIM$\uparrow$ & LPIPS$\downarrow$ \\
    \hline
    FreeNeRF \cite{yang2023freenerf}   & 3 h & 0.11 & - & {19.63} & {0.612} & {0.308} \\ 
    3DGS \cite{kerbl20233d}   & \cellcolor{red!25}{1.4 min} & 183 & \cellcolor{grey!25}{47} & 19.09 & 0.645 & {0.242} \\ 
    FSGS \cite{zhu2024fsgs}  & 12 min & {228} & 194 & \cellcolor{red!25}{20.37} & {0.696} & \cellcolor{grey!25}{0.205} \\ 
    CoRGS  \cite{zhang2024cor}  & 5 min & - & 81  &{20.24} & \cellcolor{red!25}{0.705} & \cellcolor{red!25}{0.201} \\ 

    DNGaussian \cite{li2024dngaussian}   & {3.5 min} & \cellcolor{grey!25}{240} & {78} & {20.02} & {0.682} & {0.229} \\ \hline
    Ours    & \cellcolor{grey!25}{1.7 min} & \cellcolor{red!25}{277} & \cellcolor{red!25}{36} & \cellcolor{grey!25}{20.33} & \cellcolor{grey!25}{0.699} & {0.226} \\

    \hline
    \end{tabular}}

    \label{tab:efficiency}

\end{table}

\begin{table}[!htp]
    \centering
    \caption{\textbf{Ablation of Various Initialization Methods.} The best and second-best quantitative results are marked in red and grey, respectively.}

    \begin{tabular}{l|ccccc} \hline
    Method  & FPS$\uparrow$ & \#GS(k)  & PSNR$\uparrow$ & SSIM$\uparrow$ & LPIPS$\downarrow$ \\ \hline
    3DGS w/Random      & 250 & {65} & 16.41 & 0.471 & {0.361} \\ 
    MCGS w/Random     & \cellcolor{grey!25}{260}  & \cellcolor{red!25}{33} &  19.29 & 0.638 &  0.279 \\ 
    \hline
    3DGS w/SFM     & 183 & 47 & 19.09 & 0.645 & 0.242 \\ 
    MCGS w/SFM     & 194 & {39} & {20.07} & 0.691 & 0.230 \\ \hline
    3DGS w/MVS     & 165  & 50 & 20.04 & {0.694} & \cellcolor{red!25}{0.209} \\ 
    MCGS w/MVS     & {246} & 47 & \cellcolor{red!25}{20.42} & \cellcolor{red!25}{0.708} & \cellcolor{grey!25}{0.216} \\ \hline
    3DGS w/SI      & {232} & {43} & 19.88 & 0.677 & {0.221} \\ 
    MCGS w/SI     & \cellcolor{red!25}{277} & \cellcolor{grey!25}{36} &  \cellcolor{grey!25}{20.33} & \cellcolor{grey!25}{0.699} & 0.226 \\ \hline

    \end{tabular}

    \label{tab:diffinitial}

\end{table}

\subsubsection{Ablation of Different Initializer}
To thoroughly assess the effectiveness of our sparse initializer (SI) {and the pruning strategy}, we replace the initializer with the commonly used SFM and MVS algorithms performed by COLMAP software for extremely sparse and dense initialization. We then evaluate our method (MCGS) compared to the base 3DGS architecture using these various initializers. The evaluation results are shown in Tab. \ref{tab:diffinitial}. We found that our sparse initializer significantly outperforms the COLMAP-based sparse SFM initializer on both 3DGS and our method (surpassing 0.8 dB and 0.25 dB in terms of PSNR, respectively), highlighting the advantages of our initializer. Notably, while COLMAP-based MVS initialization shows only slightly higher quantitative results, it suffers from memory inefficiency and slower rendering speeds. In contrast, the sparse initializer proposed by our method achieves a good balance between performance and efficiency. {We further perform experiments with random initialization following DNGaussian \cite{li2024dngaussian}, without relying on any geometric prior. This study demonstrates that our method maintains significant effectiveness even under extremely poor initial conditions, outperforming 3DGS by 3 dB in terms of PSNR.} Additionally, the quantitative results demonstrate the effectiveness of our MVC-guided pruning strategy across various initialization methods, with improvements of {3 dB,} 0.9 dB, 0.4 dB, and 0.5 dB in PSNR compared to 3DGS alone.

\begin{figure}[!htp]
    \centering
    \includegraphics[width=1\linewidth]{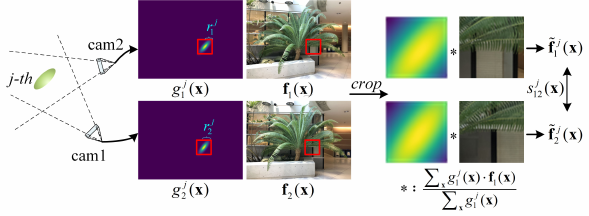}
    \caption{We present the calculation process of the feature similarity extraction considering the volume assumption. }
    \label{fig:vc}
\end{figure}

\subsubsection{Considering the Volume Assumption}
{In this paper, we sample features from the visual foundation model by simply projecting the centers of Gaussians onto the image. This approach disregards the volume assumption of 3D Gaussians and primarily discriminates against Gaussians with positional inconsistencies. Intuitively, incorporating the volume assumption—accounting for inconsistencies in position, orientation, and scale—should yield more accurate discrimination. However, we find that enforcing the volume assumption during optimization, particularly in the early training stages, actually reduces discrimination ability and introduces inefficiencies. We conduct experiments to validate this observation and demonstrate the rationale behind our simplified approach.}

{In addition to projecting the center of each Gaussian onto the imaging plane, as shown in Eq.~\ref{eq:uproject}, we also compute the corresponding 2D covariance matrix $\boldsymbol{\Sigma}^{2D}$ in the image plane, as defined by the following equation:}
\begin{equation}
    \boldsymbol{\Sigma}^{2D} = \mathbf{J}\mathbf{W}\boldsymbol{\Sigma}^{3D}\mathbf{W}^{T}\mathbf{J}^{T},
\end{equation}
{where $\mathbf{W}$ denotes the view transformation matrix, $\boldsymbol{\Sigma}^{3D}$ represents the covariance matrix of a 3D Gaussian in Euclidean space, and $\mathbf{J}$ is the Jacobian matrix of the affine approximation of the projective transformation. Given the projected mean $\boldsymbol{\mu}^j_i$ and the 2D covariance matrix $\boldsymbol{\Sigma}^j_i$ of the $j$-th Gaussian, we can compute the corresponding 2D Gaussian map as follows:}
\begin{equation}
    \mathbf{g}_i^{j}(\mathbf{x}) = \frac{1}{2\pi\sqrt{|\boldsymbol{\Sigma}^j_i|}}e^{-\frac{1}{2}(\mathbf{x}-\boldsymbol{\mu}^j_i)^{T}(\boldsymbol{\Sigma}^j_i)^{-1}(\mathbf{x}-\boldsymbol{\mu}^j_i)},
\end{equation}
{where $\mathbf{x} \in \mathbb{Z}^2$ denote the pixel coordinates. Instead of directly using the sampled feature $\mathbf{f}^j_i \in \mathbb{R}^K$ (where $i \in \{m, n\}$) from Eq. \ref{eq:sim_compute}, we replace it with an accumulated feature representation $\tilde{\mathbf{f}}^j_i \in \mathbb{R}^K$, defined as:}
\begin{equation}
    \label{eq:sim_wvolume}
    \tilde{\mathbf{f}}^j_i =  \frac{\Sigma_{\mathbf{x}\in \mathbb{A}} \mathbf{g}^j_i(\mathbf{x}) \cdot \mathbf{f}_i(\mathbf{x})}{\Sigma_{\mathbf{x}\in \mathbb{A}} \mathbf{g}^j_i(\mathbf{x})},
\end{equation}
{where $\mathbf{f}_i(\mathbf{x})$ denote the feature map of the $i^{th}$ image. Since performing feature accumulation over the entire image would be computationally inefficient and redundant, we instead compute the radius $r^j_i$ of the projected 2D Gaussian and perform accumulation only within this region $\mathbb{A}$. The radius is determined as $r^j_i = \sqrt{\lambda_{max}}$, where $\lambda_{max}$ is the largest eigenvalue of the 2D covariance matrix $\Sigma^{j}_i$.}

{To validate the effectiveness of our method under the volume assumption, we compute features using Eq. \ref{eq:sim_wvolume}, which incorporates volumetric information. The complete computational procedure is illustrated in Fig. \ref{fig:vc}. Nevertheless, for practical implementation, we adopt the simplified approach presented in Eq. \ref{eq:sim_compute}, motivated by two key considerations:}

\begin{figure}[!htp]
    \centering
    \subfigure[$it = 3000$ \& w/o VA.]{
        \includegraphics[width=0.45\linewidth]{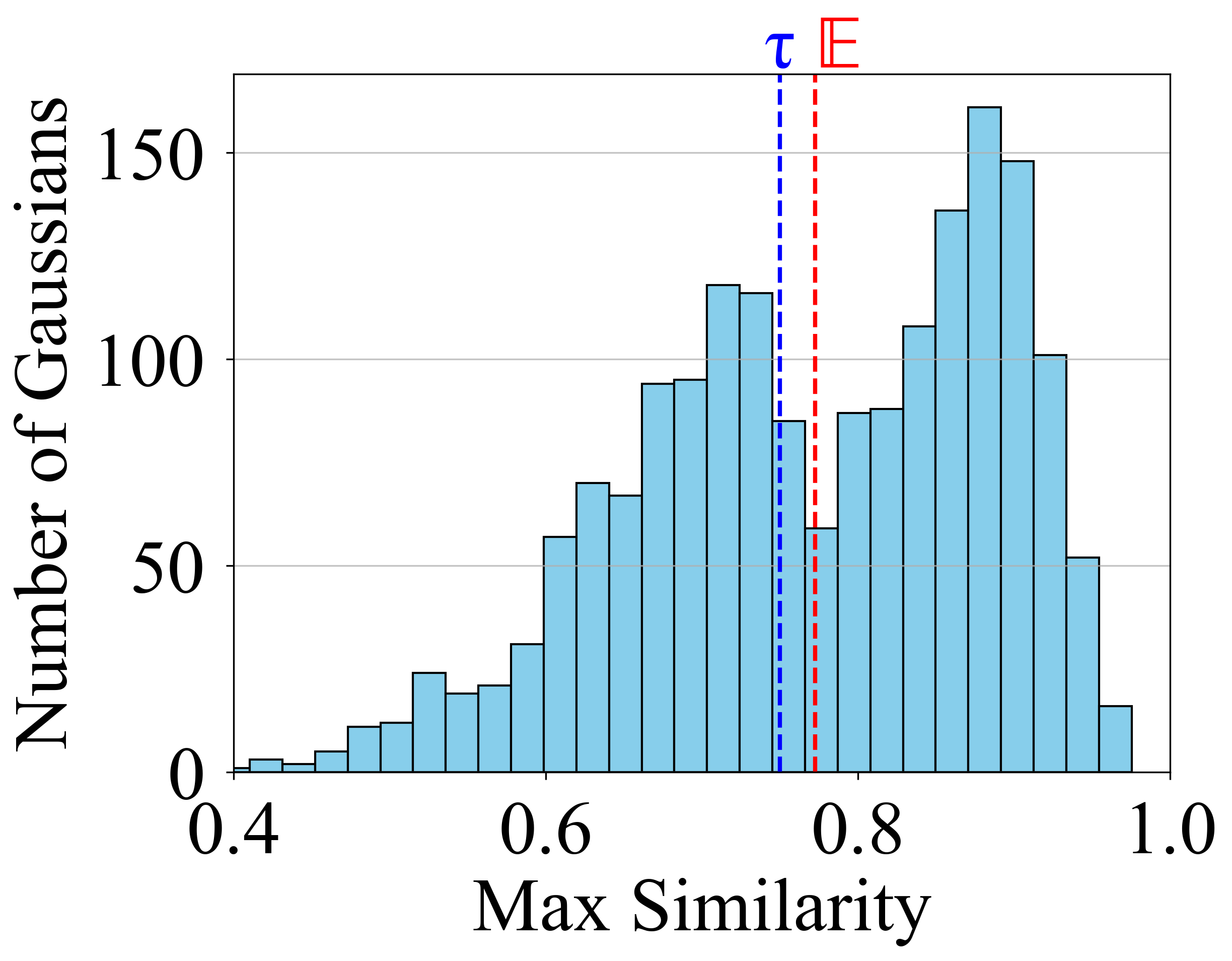}
        % \label{fig:sub1}
    }
    \hfill
    \subfigure[$it = 3000$ \& w/t VA.]{
        \includegraphics[width=0.45\linewidth]{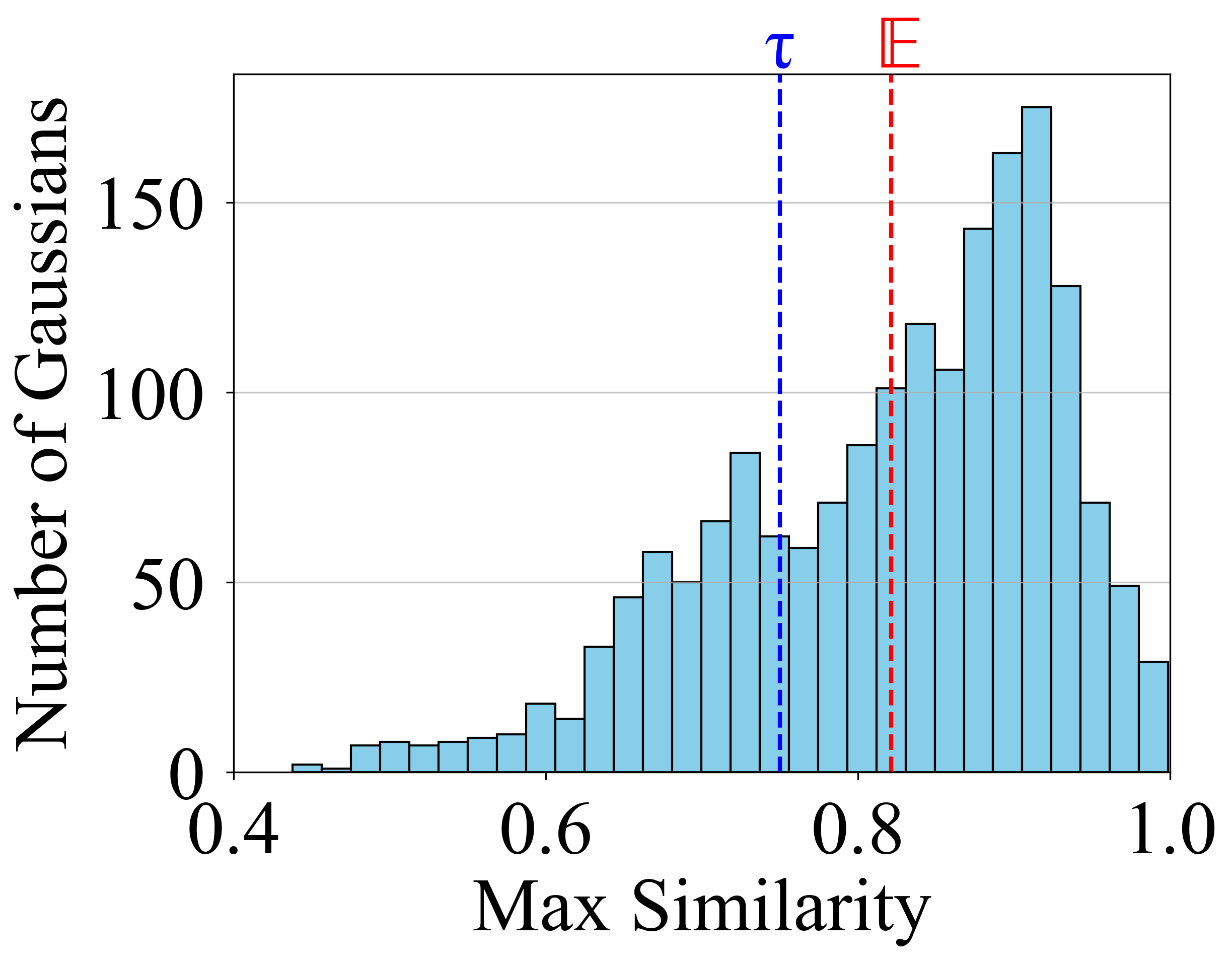}
        % \label{fig:sub2}
    }
    \hfill
    \subfigure[$it = 6000$ \& w/o VA.]{
        \includegraphics[width=0.45\linewidth]{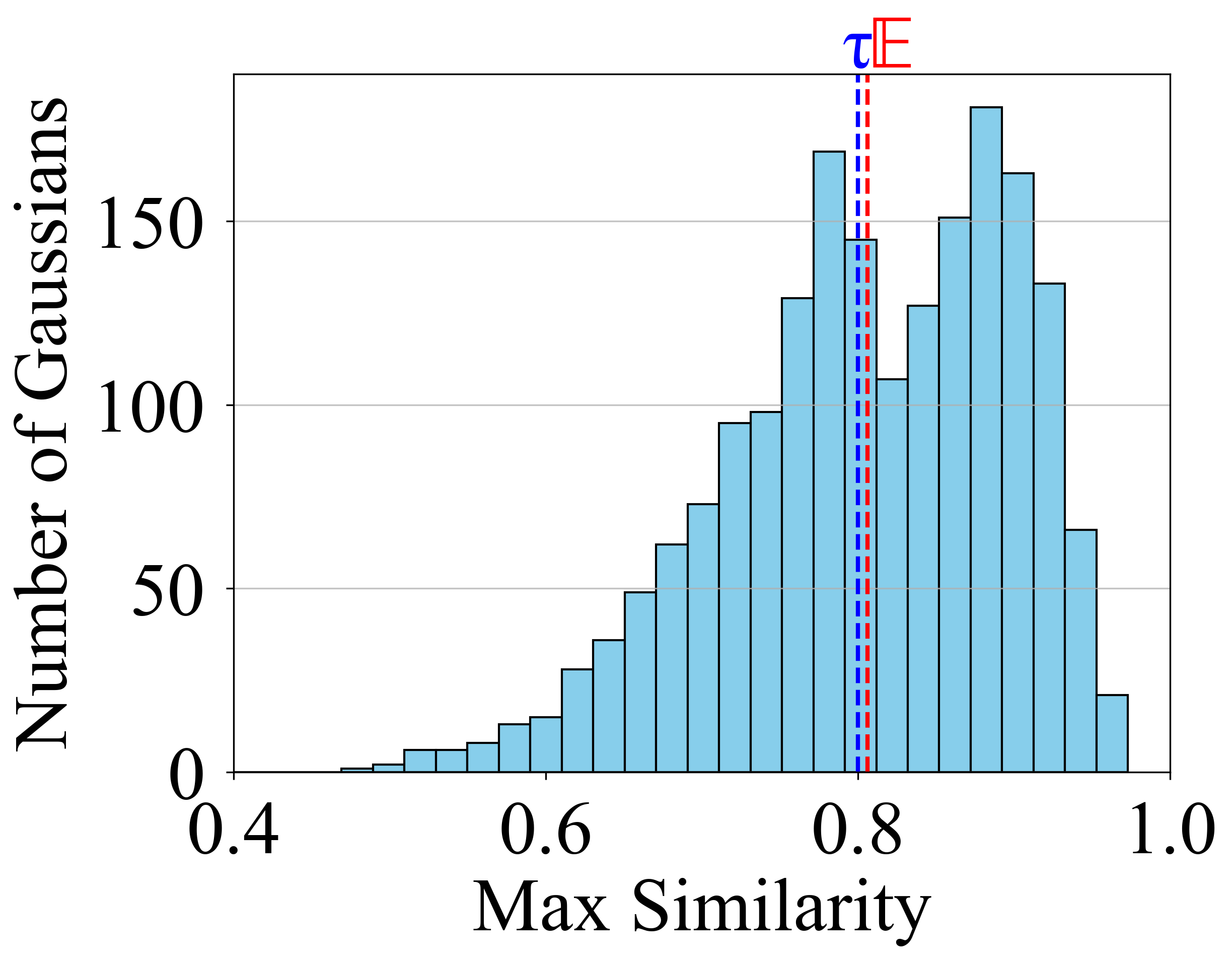}
        % \label{fig:sub3}
    }
    \hfill
    \subfigure[$it = 6000$ \& w/t VA.]{
        \includegraphics[width=0.45\linewidth]{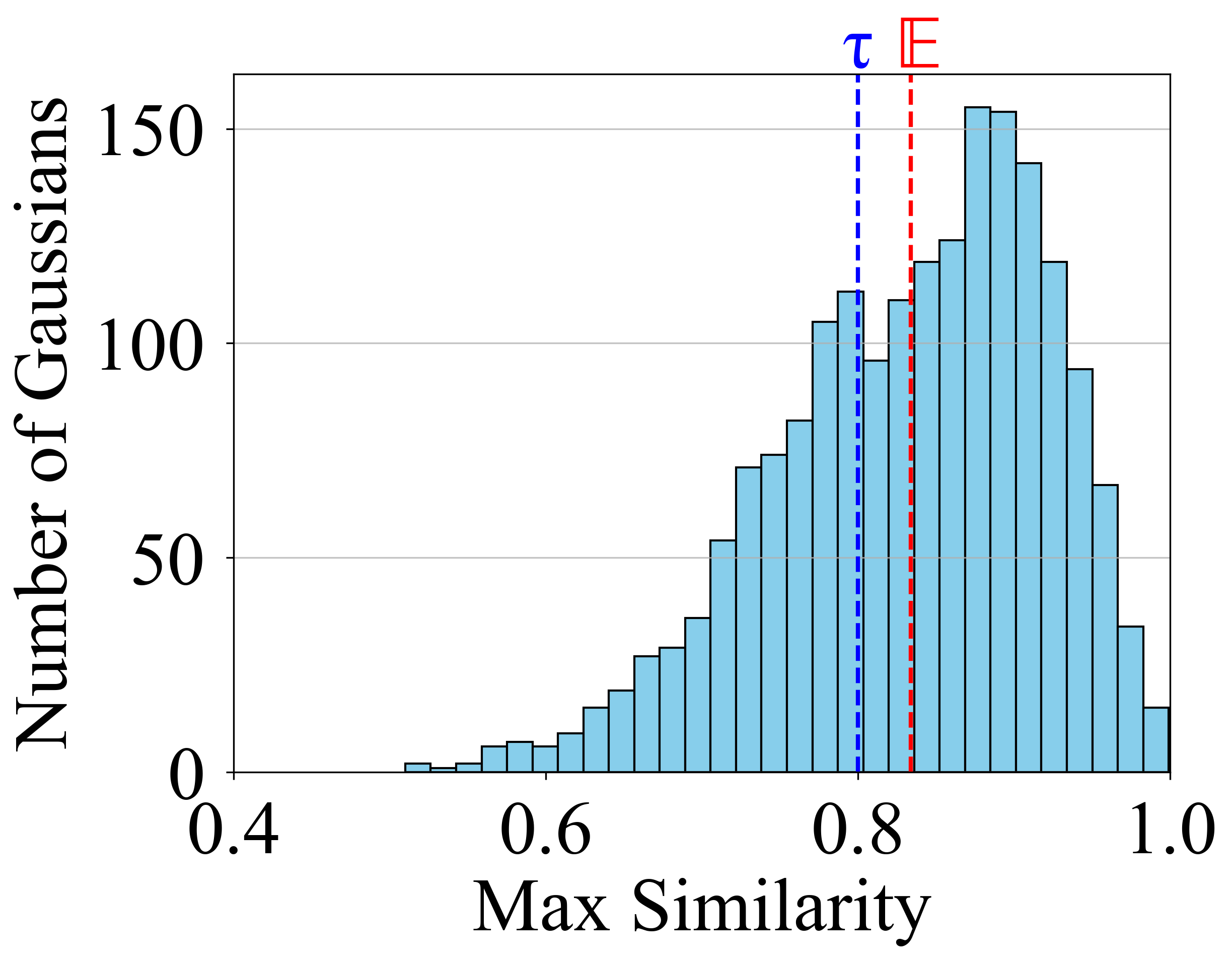}
        % \label{fig:sub4}
    }
    \hfill
    \subfigure[$it = 9000$ \& w/o VA.]{
        \includegraphics[width=0.45\linewidth]{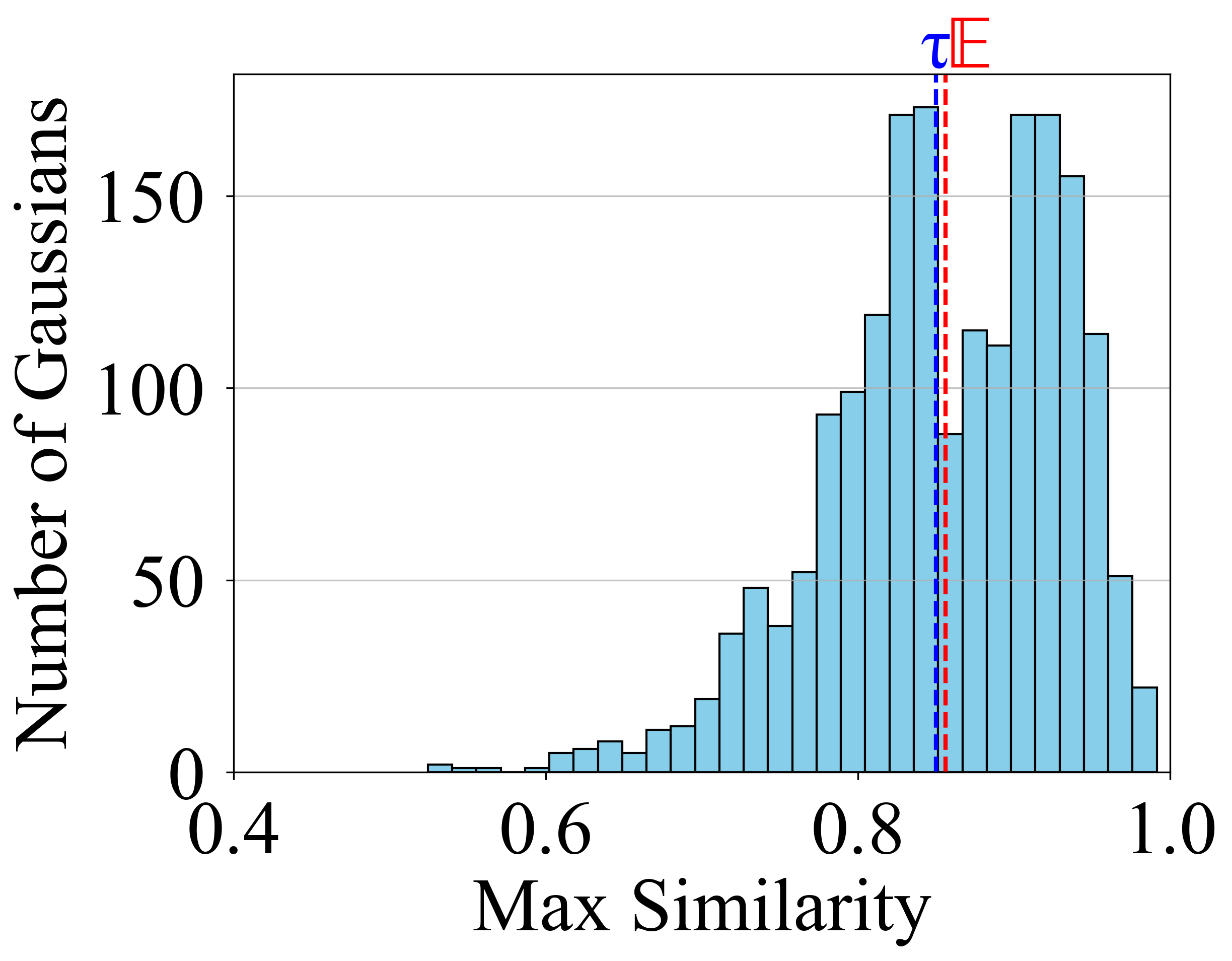}
        % \label{fig:sub3}
    }
    \hfill
    \subfigure[$it = 9000$ \& w/t VA.]{
        \includegraphics[width=0.45\linewidth]{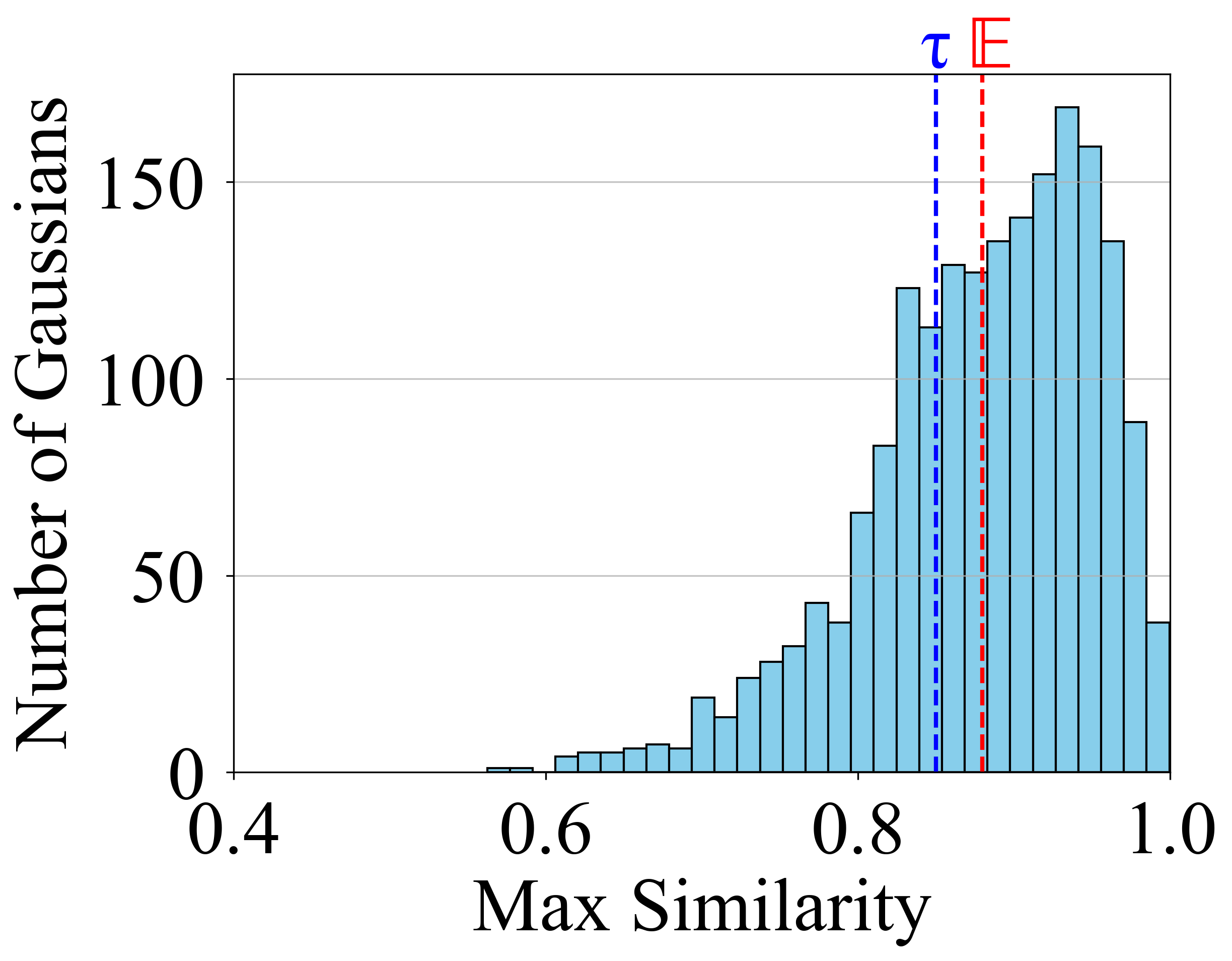}
        % \label{fig:sub4}
    }
    
    \caption{We show the maximum feature similarity distribution in each iteration step $it$ with or without considering volume assumption (VA).}
    \label{fig:distribution}
\end{figure}

\begin{table}[!htp]
    \centering
    \caption{\textbf{Quantitative Evaluation With or Without Considering Volume Assumption (VA).}}
    \label{tab:va}
    \begin{tabular}{l|cccc}\hline
        Method & Train$\downarrow$ & PSNR$\uparrow$ &  SSIM$\uparrow$ &  LPIPS$\downarrow$\\ \hline
        Ours w/t VA & \cellcolor{grey!25}{5 h} & \cellcolor{grey!25}{20.28} & \cellcolor{grey!25}{0.696} & \cellcolor{red!25}{0.220}\\ 
        Ours w/o VA & \cellcolor{red!25}{1.7 min} & \cellcolor{red!25}{20.33} & \cellcolor{red!25}{0.699} &\cellcolor{grey!25}{0.226} \\ \hline
    \end{tabular}
\end{table}

\noindent{\textbf{\textit{Discrimination Ability.}}}
{In the initial training phase, where both positional and volumetric parameters of the Gaussians remain suboptimal, incorporating volumetric considerations can impair the discriminative power of similarity metrics. This degradation arises because large Gaussian volumes, prevalent at this early stage, yield excessively broad projections onto the image plane, thereby causing significant blurring of distinctive features. Under these conditions, the Gaussian integral behaves analogously to a pooling operation applied to the feature map, effectively averaging out fine-grained details. Consequently, features that should exhibit clear distinctions based on their unique projection positions become excessively homogenized, diminishing their discriminability due to this smoothing effect.}

\begin{figure}[!htp]
    \centering
    \includegraphics[width=1\linewidth]{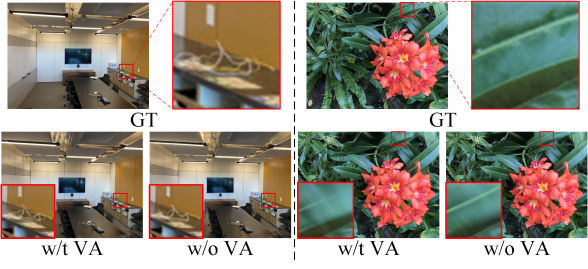}
    \caption{\textbf{Qualitative Evaluation With or Without Considering Volume Assumption (VA).}}
    \label{fig:va}
\end{figure}

{
We analyze the distribution of maximum similarity scores for Gaussians projected onto image pairs across different pruning steps (denoted by iteration counts $it$), comparing results with and without the volume assumption (VA). As illustrated in Fig. \ref{fig:distribution}, incorporating volumetric considerations leads to two notable effects: (1) an increase in the overall average similarity $\mathbb{E}$ and (2) a concentration of the distribution. These effects collectively reduce the discriminative power of the similarity metric, making optimal threshold selection more challenging for pruning. Consequently, this reduced discrimination capability may lead to the erroneous pruning of useful Gaussians, ultimately degrading model performance.
Our comprehensive evaluation includes both qualitative and quantitative assessments (Fig. \ref{fig:va} and Tab. \ref{tab:va}, respectively). The quantitative analysis reveals that similarity computation using integrated features (Eq. \ref{eq:sim_wvolume}) results in marginal decreases in both PSNR and SSIM metrics. Qualitative results corroborate these findings, demonstrating that early-stage incorporation of the volume assumption can introduce distortions in novel view synthesis.}

{Instead, as shown in Fig.~\ref{fig:distribution} (a), (c), and (e), focusing solely on positional inconsistency results in a broader distribution of similarity values, indicating a stronger discriminative capability. Notably, when only position inconsistency is considered, the distribution of maximum similarity values exhibits a clear watershed. We observe that the average similarity $\mathbb{E}$ across all Gaussians lies near the bottom of this valley, which facilitates threshold selection.
As training progresses, this watershed gradually shifts upward, suggesting that the Gaussian representations become increasingly consistent through optimization.}

\noindent{\textbf{\textit{Efficiency.}}} {During early training stages, under-optimized Gaussians with large ellipsoid scales impose significant computational overhead when performing the integration in Eq. \ref{eq:sim_wvolume}. As training progresses, the growing number of Gaussian points further exacerbates the processing burden. We quantitatively compare training times with and without the volume assumption in Tab. \ref{tab:va}. Our results demonstrate that incorporating the volume assumption leads to substantially longer training times while paradoxically yielding inferior performance—an observation that further justifies our simplified approach.}

\begin{figure*}
    \centering
    \includegraphics[width=1\linewidth]{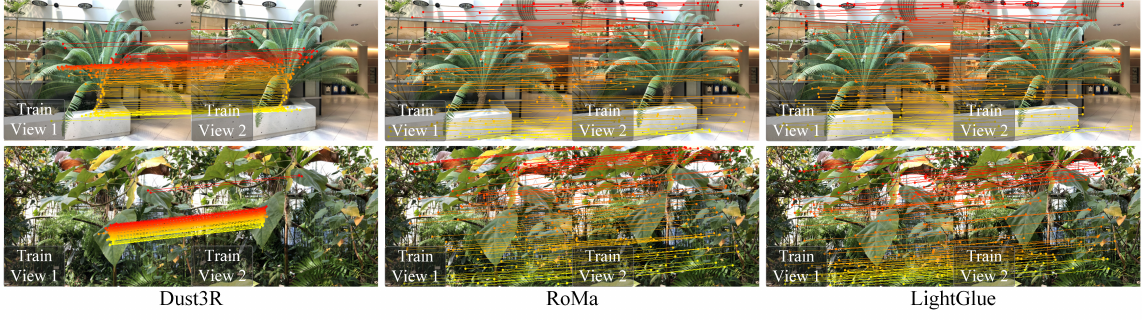}
    \caption{\textbf{Visualization of Correspondences Extracted by Different Methods.}}
    \label{fig:matches_analysis}
\end{figure*}

\subsubsection{Analysis of $\tau(t)$}
\begin{figure}[!htp]
    \centering
    \includegraphics[width=1\linewidth]{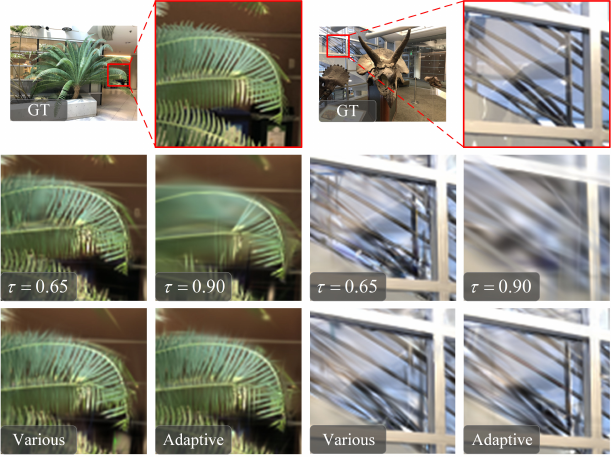}
    \caption{\textbf{Qualitative Analysis of the Threshold Parameter $\tau$ Selection.}}
    \label{fig:tau_abl}
\end{figure}

{In this work, we formulate the pruning threshold $\tau(t)$ as a dynamic parameter that evolves with the pruning step $t$. While this threshold could be treated as a fixed dataset-specific hyperparameter, we propose a more robust adaptive selection strategy motivated by our analysis of the maximum similarity distribution in Fig. \ref{fig:distribution}. Our key observation reveals a distinct bimodal distribution when considering only positional inconsistency, with the local minimum (valley) between modes approximately coinciding with the mean of maximum similarity values. Based on this finding, we implement an adaptive thresholding scheme where $\tau(t)$ is set to the average similarity value at each pruning step:}

\begin{equation}
    \label{eq:adaptive_tau}
    \tau = \mathbb{E}_j(max_{mn} (s_{mn}^{j})).
\end{equation}
{Furthermore, to account for the increasing scene consistency during optimization, we design the pruning threshold $\tau(t)$ to grow monotonically with the pruning step $t$. We systematically evaluate three thresholding strategies: (1) a constant threshold, (2) manually scheduled values, and (3) our adaptive approach based on statistical analysis (Tab. \ref{tab:abl_tau} and Fig. \ref{fig:tau_abl}). Our experiments reveal that excessively large constant thresholds (e.g., $\tau=0.9$) produce overly smoothed renderings due to excessive pruning of detail-preserving Gaussians, while overly conservative thresholds fail to effectively eliminate floaters in empty regions. Notably, our adaptive thresholding scheme (Eq. \ref{eq:adaptive_tau}) achieves comparable performance to carefully tuned manual schedules, while offering significant practical advantages: it eliminates the need for laborious parameter tuning with only marginal performance trade-offs, making it particularly suitable for real-world deployment.}

\begin{table}[!htp]
    \centering
    \caption{\textbf{Quantitative Analysis of the Threshold Parameter $\tau$ Selection.}}
    \begin{tabular}{l|c|ccc} \hline
        Strategy &  $\tau(t)$ & PSNR$\uparrow$ & SSIM$\uparrow$ &  LPIPS$\downarrow$  \\ \hline
        \multirow{6}{*}{Constant} &  [0.65, 0.65, 0.65, 0.65] & 19.99 & 0.687 &  0.221  \\ 
        &  [0.70, 0.70, 0.70, 0.70] & 20.03 & 0.692 & \cellcolor{grey!25}{0.219}   \\ 
        &  [0.75, 0.75, 0.75, 0.75] & 20.19 & 0.694 &  \cellcolor{red!25}{0.218}  \\ 
        &  [0.80, 0.80, 0.80, 0.80] & 20.12 & \cellcolor{grey!25}{0.695} &  0.222  \\ 
        &  [0.85, 0.85, 0.85, 0.85] & 20.18 & 0.692 & 0.234   \\ 
        &  [0.90, 0.90, 0.90, 0.90] & 19.55 & 0.634 & 0.304   \\  \hline
        Various &  [0.70, 0.75, 0.80, 0.85] & \cellcolor{red!25}{20.33} & \cellcolor{red!25}{0.699} & {0.226}   \\ \hline
        Adaptive &  - & \cellcolor{grey!25}{20.26}  & \cellcolor{red!25}{0.699}  & {0.223}   \\ \hline
    \end{tabular}
    
    \label{tab:abl_tau}
\end{table}

\subsubsection{Analysis of $r$}
{In Sec. \ref{sec:sparse_initial}, we introduce a constrained random filling strategy for low-texture regions. To prevent interference with matcher-initialized points, we partition the space into voxels of fixed resolution $r$ and only populate unoccupied voxels. This resolution parameter balances two competing factors: (1) minimizing perturbation to existing points and (2) maximizing filled point density. Through ablation studies (Tab. \ref{tab:r_abl}), we evaluate grid resolutions ranging from $r=16$ to $r=128$. Our results demonstrate a clear optimal range for $r$, with performance metrics peaking at intermediate values and gradually degrading as $r$ deviates from this optimum. Based on these findings, we fix $r=32$ for all experiments, as this value consistently achieves good trade-off between scene coverage and reconstruction quality.}

\begin{table}[]
    \centering
    \caption{\textbf{Quantitative Analysis of the Resolution Parameter $r$ Selection.}}
    \begin{tabular}{c|cccc} \hline
        $r$& $n_{random}$ & PSNR$\uparrow$ & SSIM$\uparrow$ &  LPIPS$\downarrow$\\ \hline
        $4$& 730 & 20.26 & 0.701 &  0.223\\
        $8$& 780 & 20.26 & 0.702 &  0.223\\
        $16$& 916 & 20.22 & 0.697 &  0.225\\
        $32$& 970 & 20.33 & 0.699 &  0.226\\
        $64$& 994 & 20.36 & 0.703 &  0.223\\ 
        $128$ & 1000 & 20.25 & 0.699 &  0.224\\ \hline
    \end{tabular}
    
    \label{tab:r_abl}
\end{table}

\subsubsection{Ablation of Various Matchers}
{In this paper, we introduce a sparse initializer that combines a sparse matching network with a random filling strategy, designed to strike a balance between rendering efficiency and reconstruction quality. Although our framework employs a specific matching network, it remains flexible and can be adapted to other methods—including both sparse and dense approaches—such as DUST3R \cite{wang2024dust3r} and RoMa \cite{edstedt2024roma}.
To thoroughly assess the generalizability and effectiveness of our approach, we conducted experiments using multiple matching techniques. Quantitative results are reported in Tab.~\ref{tab:matching}, with visualizations of the extracted correspondences shown in Figure~\ref{fig:matches_analysis}.
Our analysis indicates that correspondences derived from DUST3R’s open-source implementation are often incomplete, especially when compared to those generated by networks explicitly trained for correspondence learning—such as the sparse matcher LightGlue \cite{lindenberger2023lightglue} and the dense matcher RoMa \cite{edstedt2024roma}. This incompleteness results in weak geometric priors across many regions, significantly degrading Gaussian reconstruction quality during 3DGS initialization, as evidenced in the first row of Tab.~\ref{tab:matching}.
In contrast, our method’s random filling strategy, coupled with MVC-guided progressive pruning, effectively mitigates this issue, yielding a 2 dB improvement in PSNR. This demonstrates the robustness and effectiveness of our approach.}

{Additionally, we performed experiments using the state-of-the-art dense matcher RoMa \cite{edstedt2024roma}. While integrating RoMa yields the highest qualitative performance due to its rich and dense geometric priors, this improvement comes at the expense of slower rendering speeds and a significantly larger number of Gaussians—compared to sparse matchers that operate with a more compact set of initial Gaussians.}

\begin{table}[]
    \centering
    \caption{\textbf{Quantitative Comparison of Our Method and 3DGS Using Different Feature Matching Strategies.}}
    \label{tab:matching}
    \begin{tabular}{l|cccccc}\hline
        Method  & FPS$\uparrow$ & \#GS(k)$\downarrow$ &PSNR$\uparrow$ &  SSIM$\uparrow$ \\ \hline
        3DGS  w/t Dust3R \cite{wang2024dust3r} & 275 & 129 & 17.62  & 0.569 \\ 
        3DGS w/t RoMa \cite{edstedt2024roma}  & 210 & 54 & 20.51 & 0.726  \\ 
        3DGS  w/t LightGlue \cite{lindenberger2023lightglue}  & 219 & 44 & 19.61 & 0.667  \\ 
        Ours w/t Dust3R \cite{wang2024dust3r}  & 248 & 46 & 19.43  & 0.658  \\ 
        Ours w/t RoMa \cite{edstedt2024roma} & 263 & 40 & 20.62 & 0.728  \\ 
        Ours w/t LightGlue \cite{lindenberger2023lightglue} & 277 & 36 & 20.33 & 0.699  \\ \hline
    \end{tabular}
\end{table}

\subsection{Input Sparsity Analysis}
\begin{figure}
    \centering
    \includegraphics[width=1\linewidth]{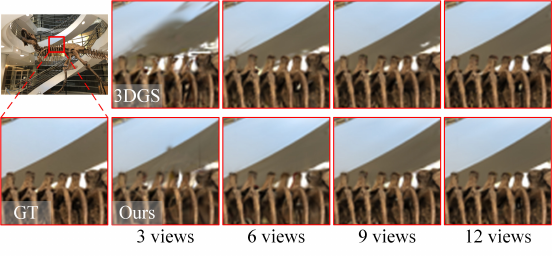}
    \caption{\textbf{Qualitative Analysis of Input Sparsity's Impact}}
    \label{fig:abl_numberview}
\end{figure}

\begin{figure}[!htp]
    \centering

    \subfigure[PSNR]{
        \includegraphics[width=0.47\linewidth]{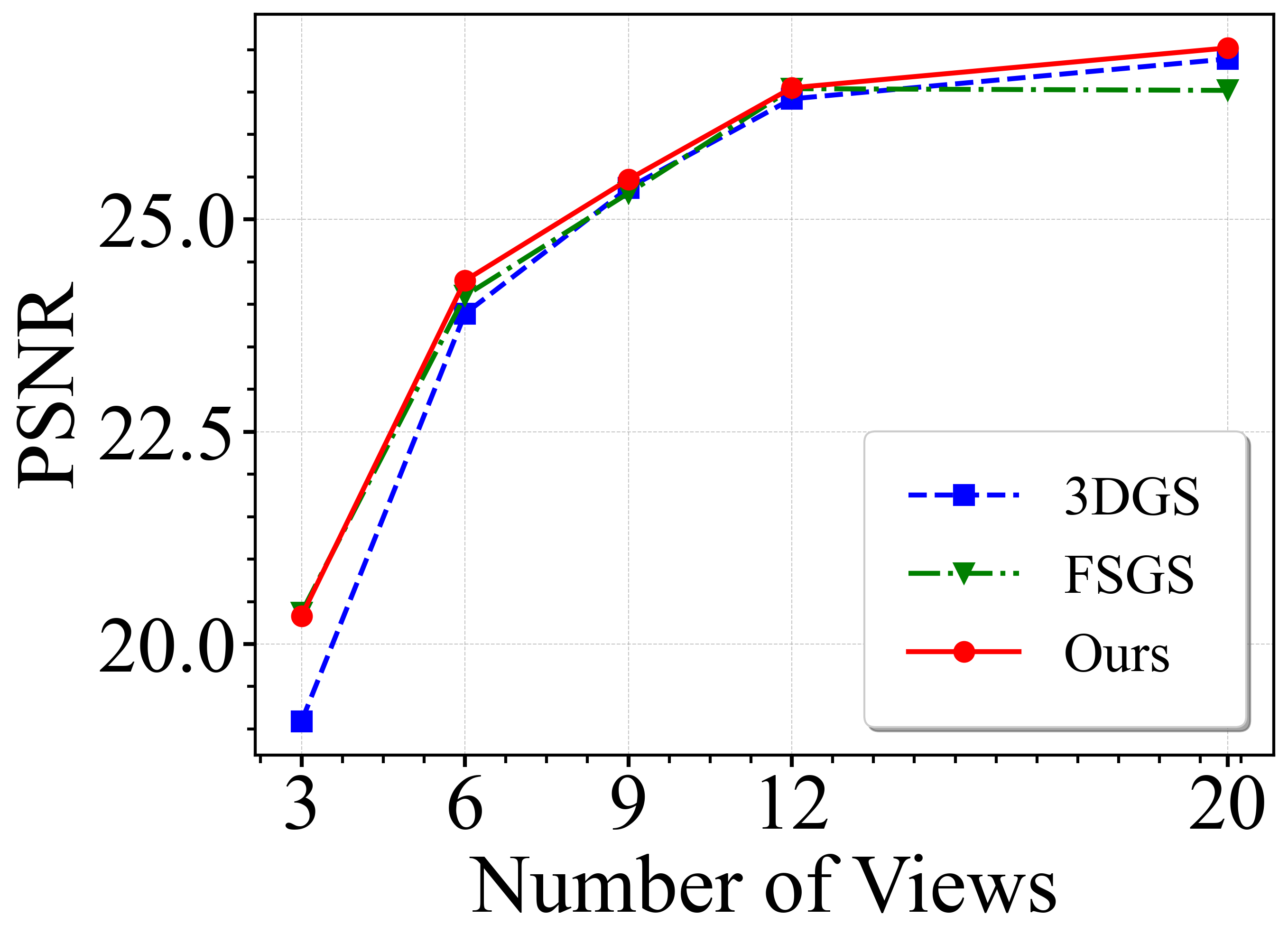}
        \label{fig:psnr_robust}
    }\subfigure[SSIM]{
        \includegraphics[width=0.46\linewidth]{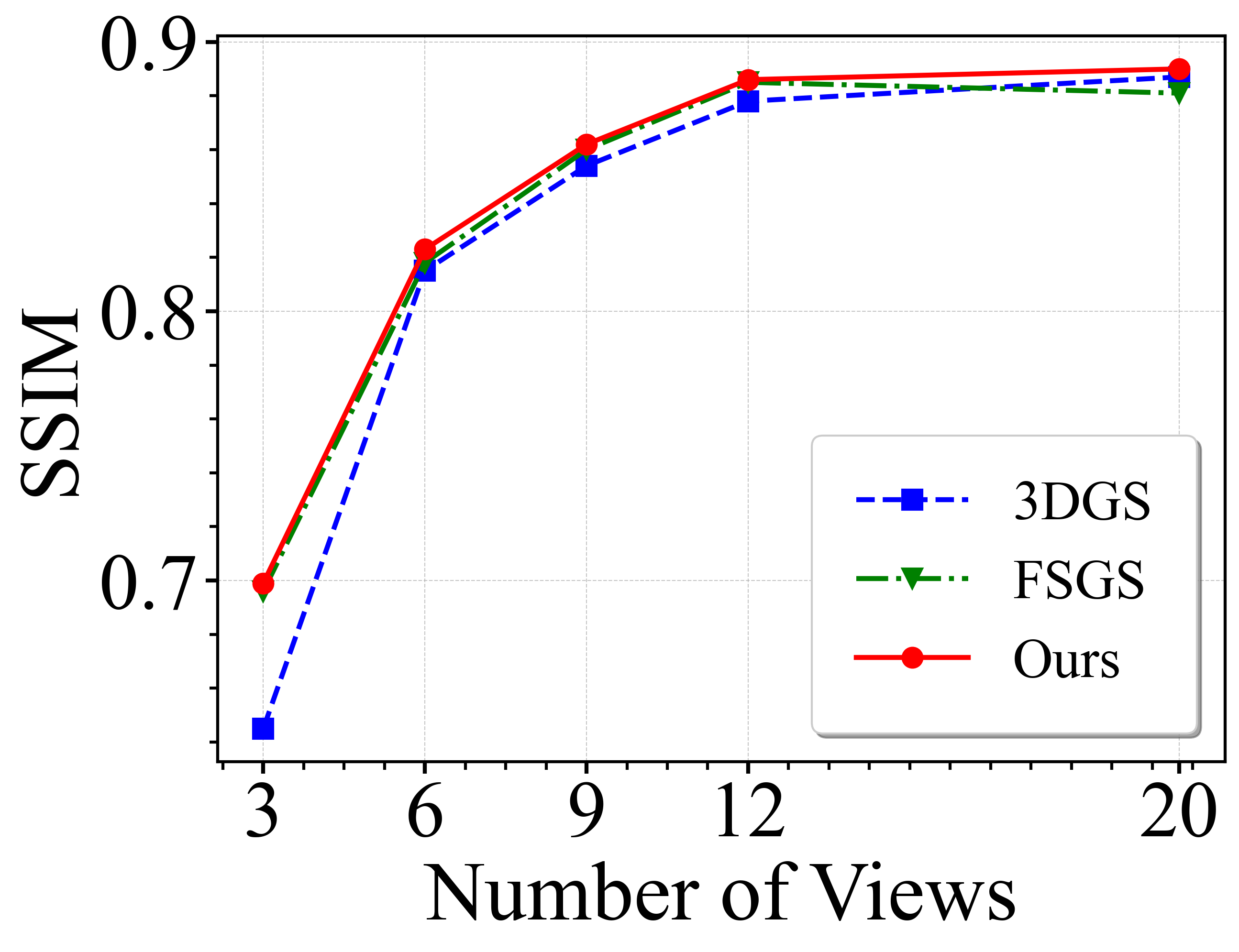}
        \label{fig:ssim_robust}
    }
    % \vskip -0.1in
    \caption{We assess the performance of our method, 3DGS, and FSGS across varying numbers of input views to evaluate their robustness to view sparsity.}
    \label{fig:robust}

\end{figure}

{To comprehensively analyze our method's dependency on input sparsity, we conducted additional experiments on the LLFF dataset. We varied the number of input views, using 3, 6, 9, 12, and 20 views, and compared our method against the baseline 3DGS and FSGS approaches.
As shown in Fig. \ref{fig:robust}, our method consistently outperforms 3DGS. While our performance is comparable to FSGS in sparse-view settings (3 to 12 views), our method achieves superior results with denser input (20 views). This demonstrates our method's robustness across various input conditions. The qualitative analysis in Fig. \ref{fig:abl_numberview} further shows that our method outperforms 3DGS in all settings, producing more accurate renderings with fewer visual artifacts. With more input views, the artifacts significantly decreased. This confirms that a sparse number of views has a significant negative impact.}

\subsubsection{Ablation Study}
\begin{figure}[!htp]

    \centering
    \includegraphics[width=1\linewidth]{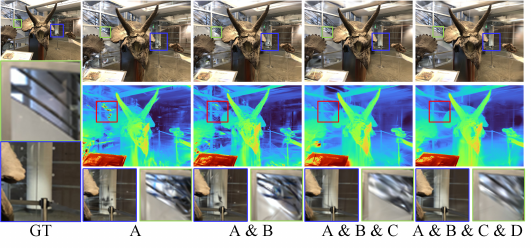}
    \caption{\textbf{Qualitative Ablation on LLFF.} We present a qualitative ablation on the LLFF dataset with 3 input views.}
    \label{fig:abl}

\end{figure}

We apply our method on the LLFF dataset using three input views. The qualitative and quantitative results of the ablation study are presented in Fig. \ref{fig:abl} and Tab. \ref{tab:abl}. The quantitative analysis shows that each component of our method—sparse initializer (A), random filling (B), multi-view consistency-guided progressive pruning (C), and edge-aware depth regularization (D)—contributes to improved performance.
Specifically, the sparse initializer (A) enhances the PSNR by 0.6 dB compared to the baseline 3DGS with the COLMAP initializer. The random filling strategy (B) further refines the initial geometry, yielding an additional 0.2 dB improvement. Multi-view consistency-guided progressive pruning (C) removes inconsistent Gaussians, resulting in a 0.4 dB increase.
The photometric and depth results in Fig. \ref{fig:abl} visually demonstrate the impact of each component. Although the edge-aware depth regularization (D) provides only minor quantitative gains, it smooths depth maps, fills geometric voids, and stabilizes the overall pipeline.

\begin{table}[!htp]

    \centering 
    \caption{\textbf{Quantitative Ablation on LLFF.}}

    \begin{tabular}{c|cccc|ccc} 
    \hline

    & A & B & C & D & PSNR$\uparrow$ & SSIM$\uparrow$ & LPIPS$\downarrow$  \\ \hline
3DGS \cite{kerbl20233d} & $\times$  & $\times$ & $\times$ & $\times$ & 19.09 & 0.645  & 0.242 \\ \hline
\multirow{3}{*}{Ours} & $\checkmark$  & $\times$ & $\times$ & $\times$ & 19.61 & 0.667 & 0.228\\
                      & $\checkmark$  & $\checkmark$ & $\times$ & $\times$ & 19.88 & 0.677 & 0.221 \\
                      & $\checkmark$  & $\checkmark$ & $\checkmark$ & $\times$ & 20.29 & 0.699 & 0.226 \\ 
                      & $\checkmark$  & $\checkmark$ & $\checkmark$ & $\checkmark$ & 20.33 & 0.699 & 0.226 \\ \hline
    \end{tabular}

    \label{tab:abl}

\end{table}

\section{Conclusion}
In this paper, we propose two novel strategies for incorporating multi-view consistency priors into 3DGS during both the initialization and optimization processes. For initialization, we introduce a sparse initializer that combines a pre-trained sparse matching network with a random filling strategy to provide sparse yet sufficient initial Gaussians. For optimization, we propose a multi-view consistency-guided progressive pruning strategy to enhance the consistency of the Gaussian field and eliminate Gaussians in empty spaces. Our method achieves state-of-the-art performance while offering faster rendering speeds and minimal memory costs.

\bibliographystyle{IEEEtran}
\bibliography{ref}

\section*{Biography Section}
\vspace{-33pt}
\begin{IEEEbiography}[{\includegraphics[width=1in,height=1.25in,clip,keepaspectratio]{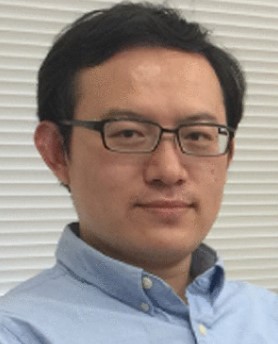}}]{Xianming Liu}
(Member, IEEE) received the BS, MS, and PhD degrees in computer science from HIT, in 2006, 2008 and 2012, respectively. He is a professor with the School of Computer Science and Technology, Harbin Institute of Technology (HIT), Harbin, China. 
In 2011, he spent half a year with the Department of Electrical and Computer Engineering, McMaster University, Canada, as a visiting student, where he then worked as a post-doctoral fellow from 2012 to 2013. He worked as a project researcher with the National Institute of Informatics (NII), Tokyo, Japan, from 2014 to 2017. He has published more than 60 international conference and journal publications, including top IEEE journals, such as IEEE Transactions on Image Processing, IEEE Transactions on Circuits and Systems for Video Technology, IEEE Transactions on Information Forensics and Security, IEEE Transactions on Multimedia, IEEE Transactions on Geoscience and Remote Sensing; and top conferences, such as CVPR, IJCAI and DCC. He is the receipt of IEEE ICME 2016 Best Student Paper Award.
\end{IEEEbiography}
\vspace{-33pt}
\begin{IEEEbiography}[{\includegraphics[width=1in,height=1.25in,clip,keepaspectratio]{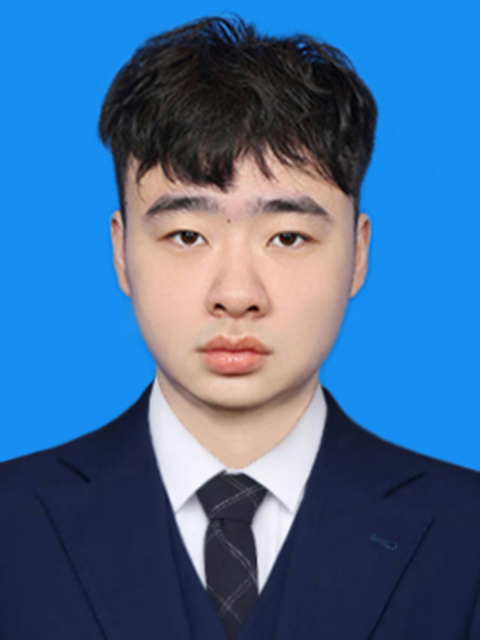}}]{Yuru Xiao}
received M.Eng. degree in 2022 from Harbin Institute of Technology, China. He is now pursuing a Ph.D. degree at the School of Computer Science and Technology, Harbin Institute of Technology. His current research focuses on 3D vision, computer graphics and neural rendering.
\end{IEEEbiography}

\vspace{-33pt}

\begin{IEEEbiography}[{\includegraphics[width=1in,height=1.25in,clip,keepaspectratio]{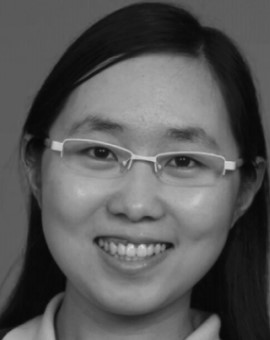}}]{Deming Zhai}
(Member, IEEE) received the B.S., M.S., and Ph.D. (Hons.) degrees in computer science from the Harbin Institute of Technology (HIT), Harbin, China, in 2007, 2009, and 2014, respectively. She is currently an Associate Professor with the Department of Computer Science, HIT. 
In 2011, she was with The Hong Kong University of Science and Technology, as a Visiting Student. In 2012, she was with the GRASP Laboratory, University of Pennsylvania, PA, USA, as a Visiting Scholar. From August 2014 to April 2016, she was a Project Researcher with the National Institute of Informatics (NII), Tokyo, Japan.
\end{IEEEbiography}
\vspace{-33pt}

\begin{IEEEbiography}[{\includegraphics[width=1in,height=1.25in,clip,keepaspectratio]{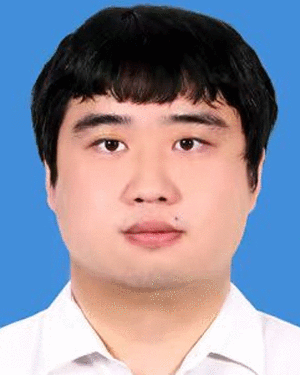}}]{Wenbo Zhao}
received the B.S., M.S., and Ph.D. degrees from Harbin Institute of Technology (HIT), Harbin, China, in 2012, 2014, and 2020, respectively. He is currently a Post-Doctor at Peng Cheng Laboratory, Shenzhen, China. His current research interests are mesh denoising, point cloud compression, and deep learning.
\end{IEEEbiography}
\vspace{-33pt}
\begin{IEEEbiography}[{\includegraphics[width=1in,height=1.25in,clip,keepaspectratio]{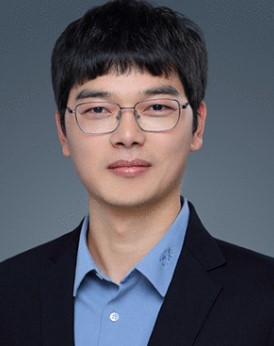}}]{Kui Jiang}
(Member, IEEE) received the Ph.D. degree from the School of Computer Science, Wuhan University, Wuhan, China, in 2022. He is currently an Associate Professor with the School of Computer Science and Technology, Harbin Institute of Technology, Harbin, China. 
His research interests include image/video processing and computer vision. He was the recipient of the 2023 CSIG Excellent Doctoral Dissertation Award and 2022 ACM Wuhan Doctoral Dissertation Award.
\end{IEEEbiography}
\vspace{-33pt}
\begin{IEEEbiography}[{\includegraphics[width=1in,height=1.25in,clip,keepaspectratio]{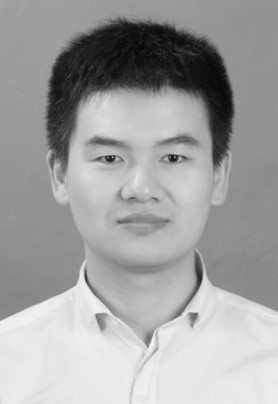}}]{Junjun Jiang}
(Senior Member, IEEE) received the B.S. degree in Information and Computing Science from Huaqiao University, Quanzhou, China, in 2009, and the Ph.D. degree in Computer Science and Technology from Wuhan University, Wuhan, China, in 2014. He is currently a Professor with the School of Computer Science and Technology, Harbin Institute of Technology, Harbin, China. 
He received the 2016 China Computer Federation (CCF) Outstanding Doctoral Dissertation Award and 2015 ACM Wuhan Doctoral Dissertation Award, His research interests include image processing and computer vision.
\end{IEEEbiography}

\vfill

\end{document}